\documentclass[sigconf]{acmart}

\usepackage{enumitem}
\usepackage{multirow}
\usepackage{tabularray}
\usepackage{tikz}
\usepackage{bbding}
\let\titleold\title
\renewcommand{\title}[1]{\titleold{#1}\newcommand{\thetitle}{#1}}
\def\maketitlesupplementary
   {
   \newpage
       \twocolumn[
        \centering
        \Large
        \textbf{\thetitle}\\
        \vspace{0.5em}Supplementary Material \\
        \vspace{1.0em}
       ] 
   }

\AtBeginDocument{%
  \providecommand\BibTeX{{%
    \normalfont B\kern-0.5em{\scshape i\kern-0.25em b}\kern-0.8em\TeX}}}

\renewcommand\footnotetextcopyrightpermission[1]{} 
\setcopyright{none}

\begin{document}

\title{Instance-free Text to Point Cloud Localization with Relative Position Awareness}

\author{Lichao Wang}
\affiliation{%
  \institution{FNii, CUHKSZ}
  \country{}
}
\email{wanglichao1999@outlook.com}

\author{Zhihao Yuan}
\affiliation{%
  \institution{FNii and SSE, CUHKSZ}
  \country{}
}
\email{zhihaoyuan@link.cuhk.edu.cn}

\author{Jinke Ren}
\affiliation{%
  \institution{FNii and SSE, CUHKSZ}
  \country{}
}
\email{jinkeren@cuhk.edu.cn}

\author{Shuguang Cui}
\affiliation{%
  \institution{SSE and FNii, CUHKSZ}
  \country{}
}
\email{shuguangcui@cuhk.edu.cn}

\author{Zhen Li}
\authornote{Corresponding author: Zhen Li (lizhen@cuhk.edu.cn).}
\affiliation{%
    \institution{SSE and FNii, CUHKSZ}
  \country{}
}
\email{lizhen@cuhk.edu.cn}



\begin{abstract}
  Text-to-point-cloud cross-modal localization is an emerging vision-language task critical for future robot-human collaboration. It seeks to localize a position from a city-scale point cloud scene based on a few natural language instructions. In this paper, we address two key limitations of existing approaches: 1) their reliance on ground-truth instances as input; and 2) their neglect of the relative positions among potential instances. Our proposed model follows a two-stage pipeline, including a coarse stage for text-cell retrieval and a fine stage for position estimation. In both stages, we introduce an instance query extractor, in which the cells are encoded by a 3D sparse convolution U-Net to generate the multi-scale point cloud features, and a set of queries iteratively attend to these features to represent instances. In the coarse stage, a row-column relative position-aware self-attention (RowColRPA) module is designed to capture the spatial relations among the instance queries. In the fine stage, a multi-modal relative position-aware cross-attention (RPCA) module is developed to fuse the text and point cloud features along with spatial relations for improving fine position estimation. Experiment results on the KITTI360Pose dataset demonstrate that our model achieves competitive performance with the state-of-the-art models without taking ground-truth instances as input.
\end{abstract}



\keywords{3D Vision and Language, text-to-point-cloud localization, Relative Position-aware Attention, 3D Transformer}

\maketitle

\section{Introduction}
\begin{figure}[h]
    \begin{center}
    \includegraphics[width=0.465\textwidth]{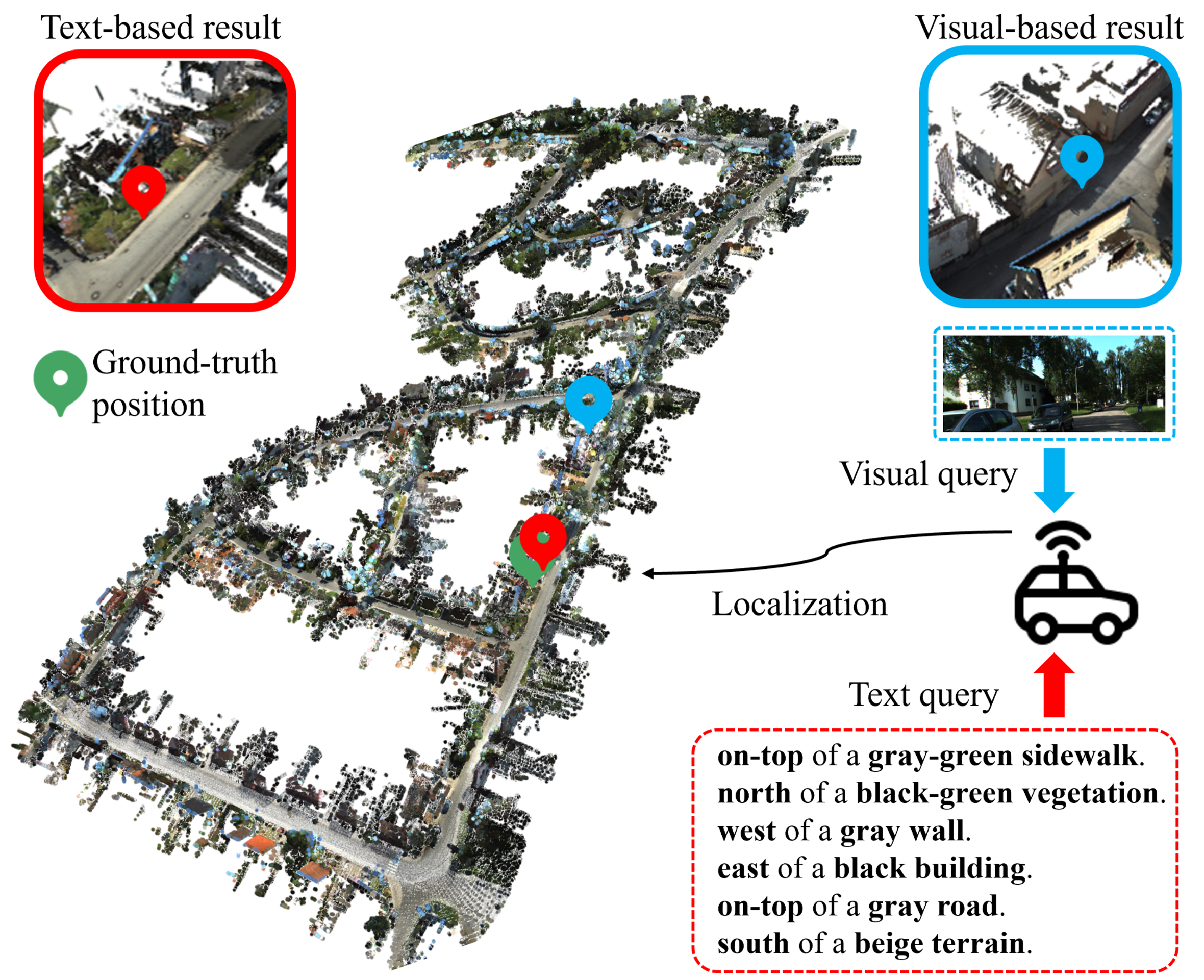}
    \end{center}
    \vspace{-1em}
    \caption{Overview of text-to-point cloud localization task. Different from visual-based place recognition task, the objective of our work is to identify a specific position within a city-scale point cloud scene based on a text query.}
    \label{fig:task_illustration}
    \vspace{-0.5em}
\end{figure}

Text-to-point-cloud localization is crucial for autonomous agents. To effectively navigate and complete tasks, future autonomous systems like self-driving vehicles and delivery drones will require collaboration and coordination with humans, necessitating the ability to plan routes and actions based on human input. Traditionally, Mobile agents employ visual localization techniques 
 \cite{Zhou_Sattler_Pollefeys_Leal-Taixe_2020,Arandjelovic_Gronat_Torii_Pajdla_Sivic_2018} to determine positions within a map, matching captured images/point clouds against databases of images/point cloud maps \cite{Arandjelovic_Gronat_Torii_Pajdla_Sivic_2018, Hausler_Garg_Xu_Milford_Fischer_2021, Xia2020SOENetAS, Sarlin_Cadena_Siegwart_Dymczyk_2019}. However, visual input is not the most efficient way for humans to interact with autonomous agents \cite{Kapoor2024OmniACTAD}. On the other hand, utilizing natural language description enhances the efficiency of communicating pick-up or delivery points to autonomous agents, as demonstrated in Fig.~\ref{fig:task_illustration}. Moreover, the text-to-point-cloud approach offers greater robustness compared to GPS systems \cite{loc_survey}. GPS systems struggle in areas concealed by tall buildings or dense foliage. In contrast, point cloud is a geometrically stable data, providing reliable localization capabilities despite environmental changes.

To accurately interpret the language descriptions and semantically understand the city-scale point clouds, a pioneer work, Text2Pos \cite{Kolmet2022Text2PosTC}, first divides a city-wide point cloud into cells and accomplishes the task using a coarse-to-fine approach. However, this method describes a cell only based on its instances, ignoring the instance relations and hint relations. To address this drawback, a relation enhanced transformer (RET) network \cite{RET} is proposed. RET captures hint relations and instance relations, and then uses cross-attention to fuse instance features into hint features for position estimation. However, RET only incorporates instance relations in the coarse stage, and the design of the relation-enhanced attention may lose crucial relative position information. More recently, Text2Loc \cite{xia2024text2loc} employes a pre-trained T5 language model \cite{T5} with a hierarchical transformer to capture contextual feature across hints, and proposes a matching-free regression model to mitigate noisy matching in fine stage. Additionally, it utilizes contrastive learning to balance positive and negative text-cell pairs. However, all current approaches rely on ground-truth instances as input, which are costly to obtain in new scenarios. While an instance segmentation model can be adopted as a prior step to provide instances, this pipeline suffers from error propagation and increased inference time, severely degrading its effectiveness. Moreover, current works do not fully leverage the relative position information of potential instances, which is crucial given that directions between the position and surrounding instances frequently appear in textual descriptions.

To address the aforementioned problems, we propose an instance query extractor that takes raw point cloud and a set of queries as input and outputs the refined quires with instance semantic information. To acquire the spatial relations, a mask module is introduced to generate the instance masks of the corresponding queries. In the coarse stage, the row-column relative position-aware self-attention (RowColRPA) mechanism is employed to capture global information of the instance queries with spatial relations. Subsequently, a max-pooling layer is employed to obtain the cell feature. In the fine stage, the instance queries undergo the relative position-aware cross-attention (RPCA) to fuse the multi-modal features with the relative position information, which effectively leverages the spatial relations and the contextual information from both point cloud and language modalities. By leveraging the instance query extractor and the relative position-aware attention mechanisms, our method addresses the two limitations, including the dependency on ground-truth instance as input and the failure to capture relative position information.

To summarize, the main contributions of this work are as follows:
\begin{itemize} [leftmargin=*]
\item We propose an instance-free relative position-aware text-to-point-cloud localization (IFRP-T2P) model, which adopts instance queries to represent potential instances in cells, thus removing the reliance on ground-truth instances as input. 
\item We design a RowColRPA module in the coarse stage and a RPCA module in the fine stage to fully leverage the spatial relation information of the potential instances.
\item We conduct extensive experiments on the KITTI360Pose \cite{Kolmet2022Text2PosTC} dataset and illustrate that our model achieves comparable performance to the state-of-the-art models without using ground-truth instances as input.
\end{itemize}

\begin{figure*}[ht]
    \begin{center}
    \includegraphics[width=0.995\textwidth]{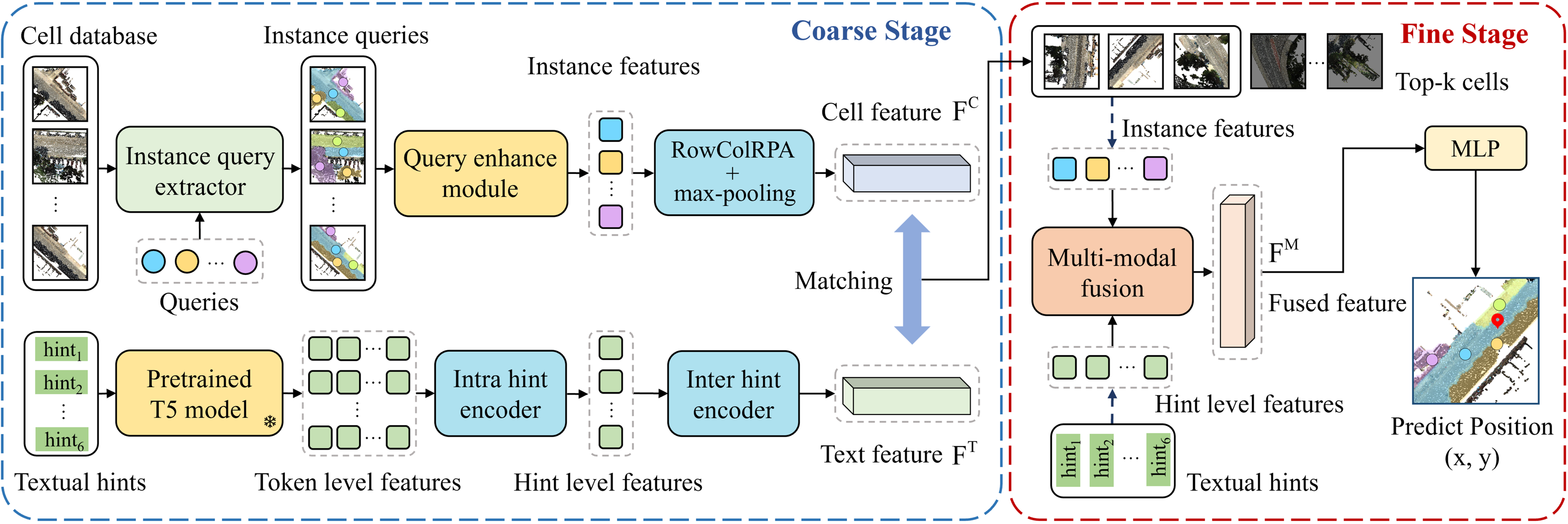}
    \end{center}
    \vspace{-1em}
    \caption{Illustration of our coarse-to-fine pipeline. Our approach processes raw point clouds of cells directly, and uses queries to represent potential instances. Firstly, the coarse stage involves selecting potential target-holding candidate cells through the retrieval of the top-$k$ cells from a pre-established cell database. Subsequently, the fine stage fuses the multi-modal features and refines the center coordinates of the selected cells.
    }
    \label{fig:pipeline}
\end{figure*}

\section{Related Work}
\subsection{3D Place Recognition}
3D place recognition aims to match a query image or LiDAR scan of a local region against a database of 3D point clouds to determine the location and orientation of the query relative to a reference map.
This technique can be categorized into two approaches, i.e., dense-point based and sparse-voxel based \cite{loc_survey}. Specifically, PointNetVLAD \cite{Uy2018PointNetVLADDP} represents a foundational work in the dense-point stream, extracting features with PointNet \cite{PointNet++} and aggregating them into a global descriptor via NetVLAD \cite{Arandjelovic_Gronat_Torii_Pajdla_Sivic_2018}. With the advent of transformer \cite{attn}, many studies \cite{Deng2018PPFNetGC, Fan_Song_Liu_Lu_He_Du_2022, Ma2023CVTNetAC, Zhou_Zhao_Adolfsson_Su_Gao_Duckett_Sun_2021} have leveraged attention mechanisms to highlight significant local features for place recognition. In contrast, voxel-based methods start by voxelizing 3D point clouds, improving the robustness and efficiency. MinkLoc3D \cite{Komorowski2020MinkLoc3DPC} extracts the local features from sparse voxelized point clouds using a feature pyramid network, followed by pooling to form a global descriptor. However, voxelization can lead to local information loss. Addressing this, CASSPR \cite{Xia_2023_ICCV} combines voxel-based and point-based advantages with a dual-branch hierarchical cross-attention transformer. Breaking away from conventional point cloud query reliance, our method employs natural language descriptions to specify target locations, offering a novel direction in 3D localization. 

\subsection{Language-based 3D Localization} 
Language-based 3D localization aims to determine a position in a city-scale 3D point cloud map based on a language description. Text2Pos \cite{Kolmet2022Text2PosTC} pioneers this field by dividing the map into cells and employing a coarse-to-fine methodology, which initially identifies approximate locations via retrieval and subsequently refines pose estimation through regression. The coarse stage matches cell features and description features in a shared embedding space, while the fine stage conducts instance-level feature matching and calculates positional offsets for final localization. However, this method overlooks the relationships among hints and point cloud instances. Addressing this gap, RET \cite{RET} incorporates relation-enhanced self-attention to analyze both hint and instance connections, and employs cross-attention in regression to improve multi-modal feature integration. Text2Loc \cite{xia2024text2loc} leverages the T5 \cite{2020t5} pre-trained language model and a hierarchical transformer for improved text embedding and introduces a matching-free fine localization approach, leveraging a cascaded cross-attention transformer for superior multi-modal fusion. By utilizing contrastive learning, Text2Loc significantly outperforms existing models in accuracy. Nonetheless, current methods depend on ground-truth instances as input, which are costly for new environments, and fail to fully leverage relative position information.

\subsection{Understanding of 3D Vision and Language}

In the realm of 3D vision and language understanding, significant strides have been made in indoor scenarios \cite{hong20233d, yan2023comprehensive, yuan2022toward, yuan2023visual}. ScanRefer \cite{chen2020scanrefer} and Referit3D \cite{achlioptas2020referit_3d} set the initial benchmarks for 3D visual grounding through the alignment of object annotations with natural language descriptions on ScanNet \cite{dai2017scannet}. Follow-up research \cite{yuan2021instancerefer, Yang_Zhang_Wang_Luo_2021, wu2022eda} improve the performance by incorporating additional features like 3D instance features, 2D prior features, and parsed linguistic features to enhance performance. 3D dense captioning \cite{chen2021scan2cap} emphasizes the detailed description of object attributes and has further been refined by integrating various 3D comprehension techniques \cite{Yuan2022XC, Cai20223DJCGAU, Chen2023EndtoEnd3D}. Conversely, the outdoor domain remains relatively untapped, with most existing datasets \cite{Malla2022DRAMAJR, Kim2018TextualEF, Nie2023Reason2DriveTI} reliant on 2D inputs. Recently, the TOD3Cap \cite{jin2024tod3cap} dataset brings 3D dense captioning into outdoor scenarios. The corresponding model leverages 2D-3D representations and combines Relation Q-Former with LLaMA-Adapter \cite{Zhang2023LLaMAAdapterEF} to produce detailed captions for objects. Different from existing works, we focus on the outdoor city-scale text-to-point-cloud localization task, aiming to bridge the gap in outdoor 3D vision and language integration.

\section{Method}

\subsection{Overview of Our Method}
The objective of text-to-point-cloud-localization is to predict the location described in a textual query within a city-scale point cloud. We manage the extensive point cloud scene by segmenting it into multiple cubic cells with same size, denoted as ${ S = {\left \{ C_{i}: i = 1,..., M \right \}}}$. Different from previous methods, in which the cell is conceptualized as a set of instances, we directly employ the colored point clouds of cells, $C \in \mathbb{R}^{n \times 6}$ of size $n$. The textual description ${T}$ is characterized by a set of hints ${ T = {\left \{ h_{j}: j = 1,..., N \right \}}}$, where each hint elucidates the direction relation between the target location and an instance. 

Following Text2Pos \cite{Kolmet2022Text2PosTC}, our method adopts a coarse-to-fine strategy, as depicted in Fig.~\ref{fig:pipeline}. Firstly, our text-cell retrieval model processes point clouds and textual cues to identify relevant cells, as detailed in Section~\ref{sec:coarse}. In the fine stage, our position estimation model directly predicts the final coordinates of the target location based on the textual hints and retrieved cells, as described in Section~\ref{sec:fine}. Notably, this approach does not necessitate ground-truth instances as input and fully exploits spatial relations in both stages. The training objective is described in Section~\ref{sec:train_objectives}.
\begin{figure*}[t]
    \begin{center}
    \includegraphics[width=0.995\textwidth]{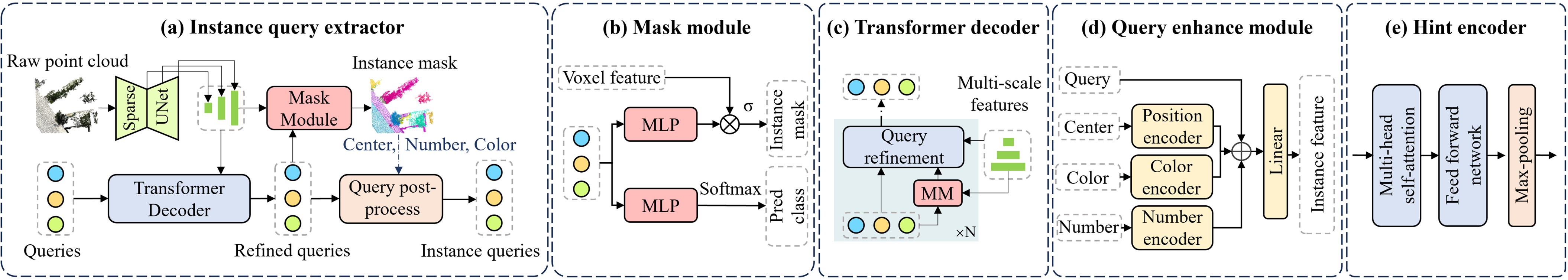}
    \end{center}
    \vspace{-1.3em}
    \caption{Illustration of (a) instance query extraction, (b) mask module, (c) transformer decoder,  (d) query enhance module, and (e) hint encoder.
    }
    \label{fig:coasre_stage}
\end{figure*}
\subsection{Coarse Text-cell Retrieval}
\label{sec:coarse}
Similar to the previous methods \cite{xia2024text2loc, Kolmet2022Text2PosTC, RET}, we employ a dual branch model to encode raw point cloud ${C}$ and text description ${T}$ into a shared embedding space, as shown in the left part of Fig.~\ref{fig:pipeline}. To directly encode the point clouds of cells, our 3D cell branch consists of three main components: the instance query extractor, the query enhance module and the row-column relative position-aware self-attention (RowColRPA) module with max-pooling layer. The instance query extractor processes the initial queries and the raw point clouds to produce instance queries and instance masks. The query enhance module merges the semantic queries with corresponding instance mask features to generate the instance features, as depicted in Fig.~\ref{fig:coasre_stage}(d). RowColRPA module with max-pooling layer, on the other hand, fuses these instance features to generate the cell feature.

\textbf{Instance query extractor.} In the instance query extractor, the raw point cloud first undergoes quantization into $\mathbf{M_0}$ voxels. Subsequently, the sparse convolution U-Net takes the voxels as input and outputs multi-scale features, achieving a maximum resolution of $\mathbf{M_0}$. For each resolution $\mathbf{r}$, we denote the feature map as $F_r$. This multi-scale features are interacted with the queries initialized by farthest
point sampling \cite{Misra21ICCV} to output the refined instance queries within the transformer decoder, as shown in Fig.\ref{fig:coasre_stage}(c). Following Mask3D \cite{Schult23ICRA}, the transformer decoder comprises $\mathbf{N}$ query refinement layers, utilizing masked cross-attention and self-attention mechanisms. Voxel feature $F_r$ is first projected to a set of keys and values, $\mathbf{K}$, $\mathbf{V}$, and the queries are projected into $\mathbf{Q}$. The following cross-attention enables queries to selectively glean information from voxel features, with mask operation specifically filtering out irrelevant voxels in the $\mathbf{Q} \cdot \mathbf{K}^{T}$ multiplication. Then, the queries undergoes a self-attention module for inter-query communication, avoiding query duplication. The instance query refinement process in each query refinement layer is outlined as:
\begin{equation}
\mathbf{Q_{ca}} = \operatorname{MMHCA}(\mathbf{Q},\mathbf{K},\mathbf{V}) + \mathbf{Q},
\end{equation}
\vspace{-1em}
\begin{equation}
\mathbf{Q_{sa}} =  \operatorname{MHSA}(\mathbf{Q_{ca}}) + \mathbf{Q_{ca}},
\end{equation}
\vspace{-1em}
\begin{equation}
\mathbf{Q}_{out} = \operatorname{FFN}
(\mathbf{Q_{sa}}) + \mathbf{Q_{sa}}, 
\end{equation}
\noindent

\noindent where $\operatorname{FFN(\cdot)}$ is the feed forward network, $\operatorname{MHSA(\cdot)}$ is the multi-head self-attention, $\operatorname{MMHCA(\cdot)}$ is the masked multi-head cross-attention. The mask module generates a binary mask for each query, as shown in Fig.~\ref{fig:coasre_stage}(b). This process involves mapping instance queries to the same feature space as the backbone feature $\mathbf{F_0}$ using an Multi-Layer Perceptron (MLP). The similarity between these mapped instance queries and $\mathbf{F_0}$ is computed through dot products, with the resulting scores undergoing a sigmoid function and subsequent thresholding at $0.5$ to yield the binary instance masks. During query post-processing, instance queries are filtered based on prediction confidence and are further combined with characteristics derived from the instance masks, resulting in the formation of the final instance queries. These instance queries are tuples that encapsulate the original queries, their center coordinates, point number, and mean RGB color value. The overall architecture and the key modules of instance query extractor are shown in Fig.~\ref{fig:coasre_stage}(a).

\begin{figure}[t]
    \vspace{-0.5em}
    \begin{center}
    \includegraphics[width=0.47\textwidth]{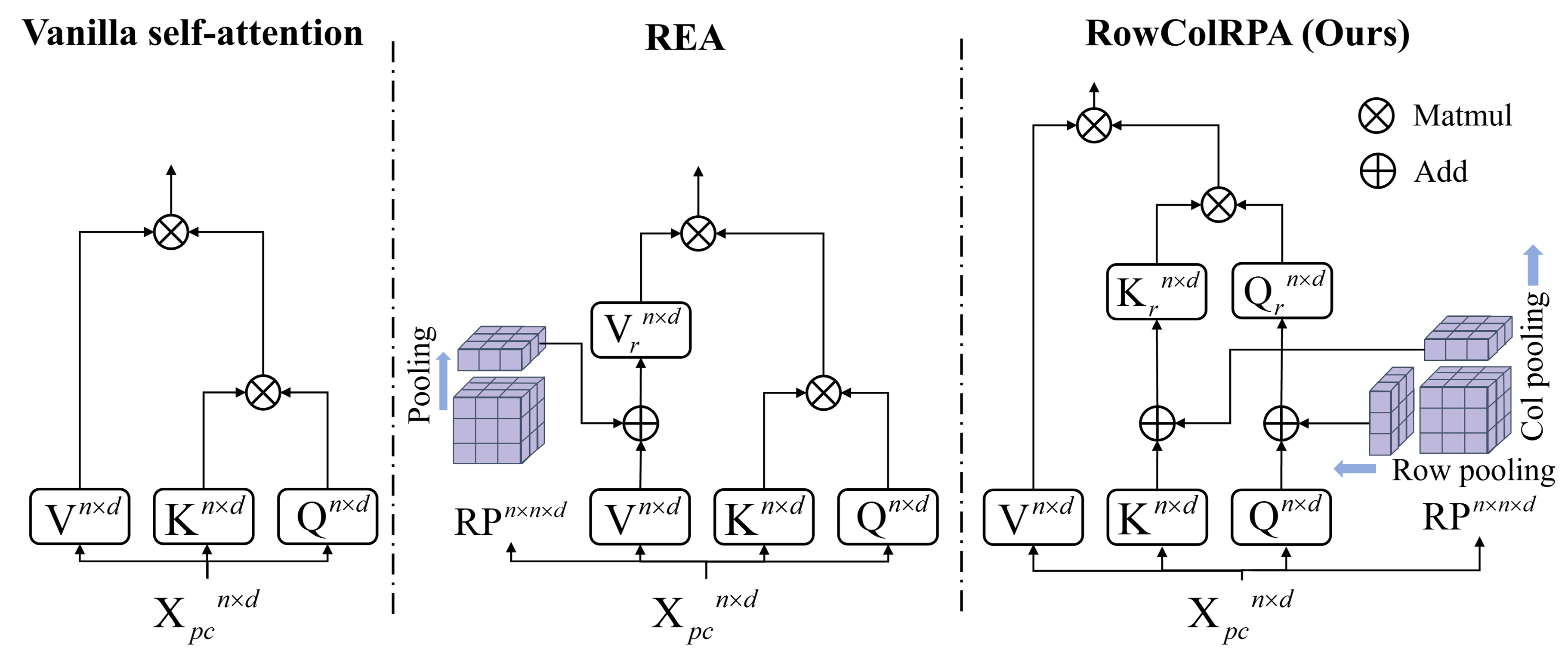}
    \end{center}
    \vspace{-1em}
    \caption{Comparison of our row-column relative position aware self-attention (RowColRPA), the relation-enhanced self-attention (REA) in RET, and vanilla self-attention.
    }
    \label{fig:row_col_rpa_comparisonpng}
\end{figure}

\textbf{Row-column relative position-aware self-attention (RowColRPA).} The RowColRPA module with max-pooling layer aims to merge instance features to produce the comprehensive cell feature $\mathbf{F^C}$, which is a common component in prior works. However, existing studies have not fully capitalized on the spatial relations among potential instances. In RET \cite{RET}, relation-enhanced self-attention (REA) loses valuable relative position information during the naive pooling and adding operation. Meanwhile, Text2Loc \cite{xia2024text2loc} focuses solely on hint relations, overlooking the relative position information of instances. To address these oversights, we introduce a RowColRPA module to exploit spatial relations in the retrieval phase, improving the distinctiveness of cell representations. Fig.~\ref{fig:row_col_rpa_comparisonpng} presents a comparison of our RowColRPA with other different self-attention mechanisms. Given the direction is often mentioned in textual queries and thus informative for retrieval, our method incorporates geometric displacement between two instances into the self-attention mechanism. The geometric displacement feature is acquired through an embedding layer as:

\vspace{-1em}
\begin{equation}
\mathbf{R} = \mathbf{W_r}(\mathbf{c}_i - \mathbf{c}_j),
\end{equation}
\vspace{-1em}

\noindent
where $\mathbf{c}_j \in \mathbb{R}^3$ represents the center coordinate of the $ith$ instance and $\mathbf{W_r} \in \mathbb{R}^{d \times 2}$ transforms the Euclidean distance into embedding space. In REA, the resulting relation embedding undergoes pooling along dimension 1 
and is subsequently added to $\mathbf{V}$. This straightforward approach, however, risks the loss of crucial relative positional information. To address this problem, we propose to fuse the relative position embedding in $\mathbf{K}$ and $\mathbf{Q}$. Therefore, during the $\mathbf{Q} \cdot \mathbf{K}^{T}$ multiplication, the spatial relation could be fully incorporated. Further, we design the row-column pooling operation to better preserve spatial relation information. The computation is described as follows:
\vspace{-1em}
\begin{equation}
\operatorname{RowColRPA}(\mathbf{Q_r}, \mathbf{K_r}, \mathbf{V}) = \operatorname{softmax}\left(\frac{\mathbf{Q_r}\mathbf{K_r}^T}{\sqrt{d_k}}\right)\mathbf{V},
\end{equation}
\vspace{-1.1em}
\begin{equation}
\mathbf{Q_r} =\mathbf{Q} + \operatorname{pooling}(\mathbf{R}, 0), 
\end{equation}
\vspace{-1.1em}
\begin{equation}
\mathbf{K_r} = \mathbf{K} + \operatorname{pooling}(\mathbf{R}, 1) ,
\end{equation}
\noindent
where $\mathbf{R} \in \mathbb{R}^{n \times n \times d}$ captures pairwise relations with $\mathbf{R}_{ij} \in  \mathbb{R}^d$ representing the relation between the $ith$ and $jth$ instance queries and $n$ representing the number of instance features. $\operatorname{Pool}(\mathbf{R}, D)$ indicates pooling tensor $\mathbf{R}$ along dimension $D \in (0, 1)$. Following the RowColRPA, a max-pooling layer is adopted to fuse the instance features into a cell feature.

In the text branch, the pre-trained T5 model and intra-inter hint encoder encode the textual hints ${h_1…h_n}$ into a comprehensive text feature $\mathbf{F^T}$. We utilize the pre-trained T5 large language model \cite{T5} for converting textual hints into embeddings at the token level. Furthermore, we utilize the intra-inter hint encoder to capture contextual relationships both within and between sentences. The intra-inter hint encoder is a hierarchical transformer, which has two transformer blocks equipped with max-pooling layers (HTM) \cite{xia2024text2loc}, as shown in Fig.~\ref{fig:coasre_stage}(e). Each HTM block can be described as follows:
\begin{equation}
\operatorname{HTM}(\mathbf{Q}, \mathbf{K}, \mathbf{V}) = \operatorname{Max-pooling} \left [ \widetilde{\mathbf{F}}_{T}+\operatorname{FFN}\left(\widetilde{\mathbf{F}}_{T}\right) \right ], 
\end{equation}

\vspace{-1em}
\begin{equation}
\widetilde{\mathbf{F}}_{T} = \mathbf{Q}+\operatorname{MHSA}\left(\mathbf{Q}, \mathbf{K}, \mathbf{V}\right),
\end{equation}
\noindent
where $\mathbf{Q} = \mathbf{K} = \mathbf{V} = \widetilde{\mathbf{F}}_{T} \in \mathbb{R}^{n_t \times d}$ represent the query, key, and value matrices, $n_t$ represents the number of hints/tokens for inter/intra hint encoder. 

\subsection{Fine Position Estimation}
\label{sec:fine}
Following the coarse stage, we aim to refine the location prediction within the retrieved cells. Based on the matching-free network \cite{xia2024text2loc}, we introduce the query instance extractor to mitigate the dependency on ground-truth instances as input. Moreover, we propose a relative position-aware cross-attention module to incorporate spatial relation information in the position estimation process. 

\begin{figure}[t]
    \begin{center}
    \includegraphics[width=0.475\textwidth]{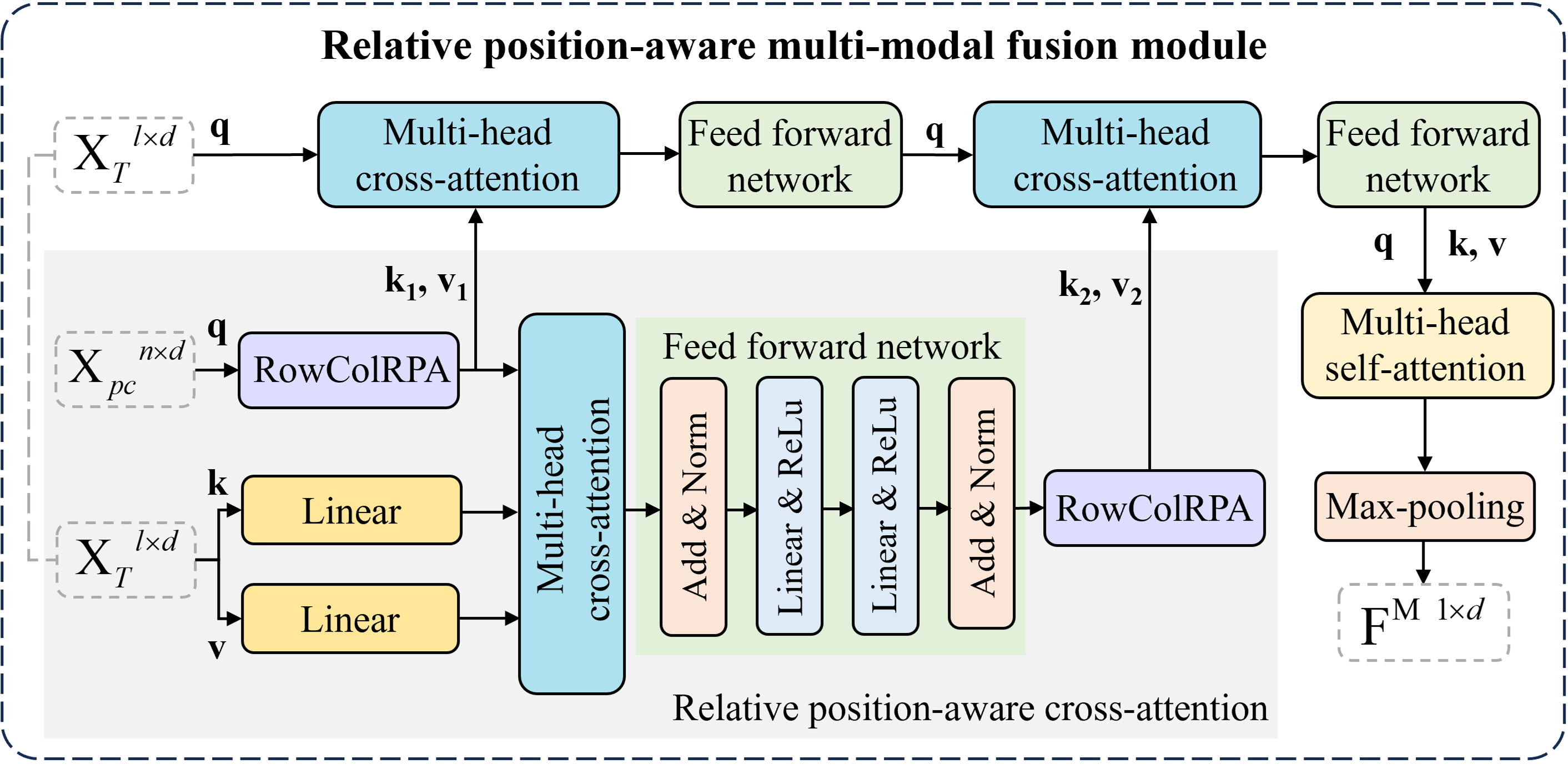}
    \end{center}
    \vspace{-0.6em}
    \caption{Illustration of the relative position-aware multi-modal fusion module. The relative-position-aware cross attention (RPCA) merges potential instance features with text keys and values, infusing semantic and spatial relation information with text embeddings.
    }
    \vspace{-0.5em}
    \label{fig:fine_stage}
\end{figure}

Similar to the coarse stage, our fine position estimation model comprises two branches. The text branch utilizes a pre-trained T5 model \cite{T5} to extract token-level features, followed by an intra hint encoder to encode the hint features. The intra hint encoder is a HTM block, as described in Section~\ref{sec:coarse}. In the point cloud branch, the instance query extractor first generates the instance queries from the raw point cloud, which are then enhanced by the corresponding instance mask features, consistent with the cell retrieval stage. Subsequently, the text-cell feature and the spatial relation information are merged into a $\mathbf{F^M}$ by the relative position-aware multi-modal fusion module, as depicted in Fig.~\ref{fig:fine_stage}. Finally, the target position is predicted via an MLP. The overall architecture of our fine stage model is depicted in the right part of Fig.~\ref{fig:pipeline}.

\textbf{Relative position-aware cross-attention (RPCA).} Within the multi-modal fusion module, we introduce the RPCA to integrate spatial relation information with the text-cell features, as shown in Fig.~\ref{fig:fine_stage}. The multi-modal fusion module consists of two cross-attention modules and one RPCA module. The two cross-attention modules are configured with the text feature serving as the query and the point cloud feature as both key and value. The RPCA takes the point cloud feature as query and the text feature as key and value. To prepare the point cloud feature for RPCA, it first passes through a RowColRPA module, which crafts the query for RPCA and also generates the key and value ($\mathbf{k_1}$, $\mathbf{v_1}$) for the first cross-attention. Concurrently, the text feature undergoes two linear layers to get the key and value. Following the cross-attention operation, an additional RowColRPA is applied to create the key and value ($\mathbf{k_2}$, $\mathbf{v_2}$) for the second cross-attention. This design incorporates spatial relation features within the multi-modal fusion process.

\subsection{Training Objectives}
\label{sec:train_objectives}
\noindent\textbf{Instance query extractor.} To ensure the effective generation of instance queries across both stages, we incorporate the instance segmentation task as an auxiliary task. To align these predicted instance masks $m$ with their accurate labels $\hat{m}$, we employ bipartite graph matching \cite{Carion20ECCV,Cheng21NeurIPS,Cheng22CVPR} to form a cost matrix for calculating matching costs, similar to the method used in Mask3D \cite{Schult23ICRA}. The Hungarian algorithm \cite{Kuhn55hungarian} is utilized to address this cost assignment problem. After establishing the correspondences, the optimization of each mask prediction is formulated as follows:
\begin{equation} \small
\mathcal{L}_{\text{mask}}(m, \hat{m}) = \mathcal{L}_{\text{BCE}}(m, \hat{m}) + \mathcal{L}_{\text{dice}}(m, \hat{m}),
\label{eq:loss_mask}
\end{equation}
where $\mathcal{L}_{\text{BCE}}$ denotes the binary cross-entropy loss and $\mathcal{L}_{\text{dice}}$ represents the Dice loss \cite{Deng18ECCV}.
To supervise the classification, we employ the default multi-class cross-entropy loss, $\mathcal{L}_{\text{CE}}$.
The overall loss function for auxiliary instance mask prediction is defined by:
\begin{equation} \small
\mathcal{L}_{\text{inst}} = \mathcal{L}_{\text{mask}}(m, \hat{m})  + \lambda_{\text{c}} \mathcal{L}_{\text{CE}}(c, \hat{c}),
\end{equation}
where $c$ and $\hat{c}$ denote the predicted class and the ground-truth class label, respectively. 

{ \bf Coarse text-cell retrieval.} We train the text-cell retieval model with a cross-modal contrastive learning objective. Given an input batch of 3D cell features $\{{F}^C_i\}^{N}_{i=1}$ and matching text description features $\{{F}^T_i\}^{N}_{i=1}$ with $N$ representing the batch size, the contrastive loss among each pair $l_{\text{CL}}(F^T_i,F^C_{i})$ is computed as follows,
\begin{equation}\small
      l_{\text{CL}}(F^T_i,F^C_{i}) = -\log\frac{\exp(F^T_i\cdot F^C_{i}/\tau)}{\sum\limits_{j\in N} \exp(F_i^T\cdot F_j^C/\tau)} - \log\frac{\exp(F^C_i\cdot F^T_{i}/\tau)}{\sum\limits_{j\in N} \exp(F_i^C\cdot F_j^T/\tau)} ,
\end{equation}
\noindent
where $\tau$ is the temperature coefficient, similar to CLIP~\cite{radford2021learning}. Within a training batch, the text-cell contrastive loss $\mathcal{L}_{\text{CL}}(F^T,F^C)$ can be described as:
\begin{equation}\small
       \mathcal{L}_{\text{CL}}(F^T,F^C) = \frac{1}{N}\left [ \sum_{i\in N} l_{\text{CL}}(F^T_i,F^C_{i}) \right ].
\end{equation}
\noindent
Combining with the auxiliary instance segmentation loss, the final coarse text-cell retrieval loss $\mathcal{L}_{\text{coarse}}$ is depicted as:
\begin{equation}\small
      \mathcal{L}_{\text{coarse}} = \mathcal{L}_{\text{CL}}(F^T,F^C) + \lambda_{\text{inst}}\mathcal{L}_{\text{inst}}.
\end{equation}
\noindent

{ \bf Fine position estimation.} In the fine stage, we aim to minimize the Euclidean distance between the predicted position and the ground-truth. In this paper, we adopt the mean squared error loss $\mathcal{L}_{\text{MSE}}$ to train the regression model, as:
\begin{equation} 
\begin{aligned}
\mathcal{L}_{\text{MSE}}(P, \hat{P}) = \big \|P - \hat{P}   \big \|_{2},
\end{aligned}
\label{Eq: regression loss}
\end{equation}
\noindent 
where $P=(x, y)$ is the predicted target coordinates and $\hat{P}$ is the corresponding ground-truth coordinates. We combine the auxiliary instance segmentation loss with the mean square error loss in the final fine position estimation loss $\mathcal{L}_{\text{fine}}$, which is define as:
\begin{equation}\small
      \mathcal{L}_{\text{fine}} = \mathcal{L}_{\text{MSE}}(P, \hat{P}) + \lambda_{\text{inst}}\mathcal{L}_{\text{inst}}.
\end{equation}
\noindent

\section{Experiments}
\begin{table*}[t]
    \centering
    \begin{tblr}{
      cells = {c},
      cell{1}{3} = {c=6}{},
      cell{2}{3} = {c=3}{},
      cell{2}{6} = {c=3}{},
      vline{6} = {1-3}{solid}, 
      hline{1,7} = {-}{0.08em},
      hline{2-3} = {3-8}{0.03em},
      hline{4,7} = {-}{0.05em},
      hline{11} = {-}{0.05em}, 
      hline{12} = {-}{0.08em}, 
    }
                    & &  Localization Recall ($\epsilon < 5/10/15m $) $\uparrow$  &   &   &   &   & \\
    Methods         & GT & Validation Set &   &   & Test Set &   &   \\
                    &   & $k=1$ & $k=5$ & $k=10$ & $k=1$ & $k=5$ & $k=10$ \\
    Text2Pos  ~\cite{Kolmet2022Text2PosTC} &   \CheckmarkBold     & 0.14/0.25/0.31                                 & 0.36/0.55/0.61                           & 0.48/0.68/0.74                            & 0.13/0.21/0.25                            & 0.33/0.48/0.52                            & 0.43/0.61/0.65                            \\
    RET  ~\cite{RET} &     \CheckmarkBold       & 0.19/0.30/0.37                                          & 0.44/0.62/0.67                            & 0.52/0.72/0.78                            & 0.16/0.25/0.29                            & 0.35/0.51/0.56                            & 0.46/0.65/0.71                            \\
    Text2Loc  ~\cite{xia2024text2loc}  &    \CheckmarkBold   & \underline{0.28/0.50/0.58}                                          & \underline{0.58/0.80/0.83}
                           & \underline{0.69/0.88/0.91}                            & \underline{0.26/0.43/0.48}                            & \underline{0.51/0.72/0.76}                            & \underline{0.60/0.81/0.85}
                           \\
    NetVlad  ~\cite{Arandjelovic_Gronat_Torii_Pajdla_Sivic_2018}  &    \XSolidBrush   & 0.18/0.33/0.43                                          & 0.29/0.50/0.61                           & 0.34/0.59/0.69                            &-                            & -                            & -                            \\
    PointNetVlad  ~\cite{Uy2018PointNetVLADDP}   &   \XSolidBrush   & 0.21/0.28/0.30                                          & 0.44/0.58/0.61                            & 0.54/0.71/0.74                            & 0.13/0.17/0.18                            & 0.28/0.37/0.39                            & 0.36/0.48/0.50                            \\
    Text2Pos  ~\cite{Kolmet2022Text2PosTC}  &    \XSolidBrush   & 0.11/0.25/0.32                                          & 0.30/0.53/0.62                           & 0.38/0.66/0.74                            & 0.09/0.20/0.24                            & 0.24/0.44/0.51                            & 0.32/0.57/0.64                            \\
    Text2Loc  ~\cite{xia2024text2loc}   &   \XSolidBrush   & 0.18/0.35/0.44                                          & 0.43/0.66/0.73                            & 0.54/0.77/0.82                            & 0.17/0.32/0.37                            & 0.40/0.59/0.64                            & 0.49/0.70/0.75                            \\
    IFRP-T2P  (Ours)  &   \XSolidBrush    & \textbf{0.23}/\textbf{0.45}/\textbf{0.53}                                          & \textbf{0.53}/\textbf{0.70}/\textbf{0.81}                            & \textbf{0.64}/\textbf{0.86}/\textbf{0.89}                            & \textbf{0.22}/\textbf{0.40}/\textbf{0.46}                            & \textbf{0.47}/\textbf{0.68}/\textbf{0.73}                            & \textbf{0.58}/\textbf{0.78}/\textbf{0.82}

    \end{tblr}
    \caption{Performance comparison on the KITTI360Pose dataset. First three methods utilize ground-truth instances as input. Other methods take raw point cloud data directly.
    }
    \label{tab:fine_results}
    \vspace{-0.3em}
\end{table*}

\begin{table}[t]
    \centering
    \resizebox{0.475\textwidth}{!}{
    \begin{tblr}{
      cells = {c},
      cell{1}{3} = {c=6}{},
      cell{2}{3} = {c=3}{},
      cell{2}{6} = {c=3}{},
      vline{6} = {1-3}{solid}, 
      hline{1,7} = {-}{0.08em},
      hline{2-3} = {3-8}{0.03em},
      hline{4,7} = {-}{0.05em},
      hline{9} = {-}{0.05em}, 
      hline{10} = {-}{0.08em}, 
    }
                    &   &      Text-cell Retrieval Recall $\uparrow$         &               &               &               &               \\
    Methods       & GT & Validation Set                     &               &               & Test Set      &               &               \\
                   &  & $k=1$                              & $k=3$         & $k=5$         & $k=1$         & $k=3$         & $k=5$         \\
    Text2Pos ~\cite{Kolmet2022Text2PosTC}  &   \CheckmarkBold    & 0.14                               & 0.28          & 0.37          & 0.12          & 0.25          & 0.33          \\
    RET ~\cite{RET}    &    \CheckmarkBold      & 0.18                               & 0.34          & 0.44          & -             & -             & -             \\
    Text2Loc  ~\cite{xia2024text2loc} &  \CheckmarkBold & \underline{0.29}                     & \underline{0.54} & \underline{0.63} & \underline{0.27} & \underline{0.47} & \underline{0.56} \\
    Text2Pos  ~\cite{Kolmet2022Text2PosTC}    &  \XSolidBrush   & 0.09                               & 0.20          & 0.26          & 0.07          & 0.18          & 0.23 \\
    Text2Loc  ~\cite{xia2024text2loc}    &  \XSolidBrush   & 0.19                               & 0.37         & 0.46          & 0.18          & 0.34          & 0.42 \\
    IFRP-T2P  (Ours)    &  \XSolidBrush  & \textbf{0.24}                      & \textbf{0.46} & \textbf{0.57} & \textbf{0.23} & \textbf{0.39} & \textbf{0.48} 
    \end{tblr}}
    \caption{Performance comparison for coarse text-cell retrieval on the KITTI360Pose dataset.
    }
    \label{tab:coarse_results}
    \vspace{-0.3em}
\end{table}

\begin{table}[thbp]
    \centering
    \scriptsize
    \setlength{\tabcolsep}{10pt} 
    \resizebox{0.475\textwidth}{!}{
    \begin{tabular}{lcc}
     \toprule[0.08em]
    Methods              & Validation Error $\downarrow$ & Test Error $\downarrow$ \\
    \midrule[0.05em]
    Text2Pos \cite{Kolmet2022Text2PosTC}           & 0.130 & 0.131             \\
    Text2Loc  \cite{xia2024text2loc}    & 0.121 &  0.122            \\
    \midrule[0.05em]
    IFRP-T2P (Ours)                    & \textbf{0.118} &  \textbf{0.119}            \\
    \bottomrule[0.08em]
    \end{tabular}
    }
    \caption{Performance comparison of the fine-stage models on the KITTI360Pose dataset. Normalized Euclidean distance is adopted as the metric. All methods directly take raw point cloud as input.}\label{tab: regressor}
    \vspace{-1em}
\end{table}

\subsection{Dataset Description}
We evaluate our IFRP-T2P model on the KITTI360Pose dataset \cite{Kolmet2022Text2PosTC}. This dataset encompasses 3D point cloud scenes from nine urban areas, spanning a city-scale space of 15.51 square kilometers and consisting of 43,381 paired descriptions and positions. We utilize five areas for our training set and one for validation. The remaining three areas are used for testing. Each 3D cell within this dataset is represented by a cube measuring 30 meters on each side, with a 10-meter stride between each cell. More details are provided in the supplementary material.

\subsection{Implementation Details}
Our approach begins with pre-training the instance query extractor. This is accomplished through the instance segmentation objective, as outlined in Section 3.2. 
Our feature backbone employs a sparse convolution U-Net, specifically a Minkowski Res16UNet34C \cite{choy20194d} with 0.15 meter voxel size. To align with the requirements of the KITTI360Pose configuration, we utilize cell point clouds, each encompassing a 30 meter cubic space, as our input. The training extends over 300 epochs, utilizing the AdamW optimizer. 
In the coarse stage, we train the text-cell retrieval model with AdamW optimizer with a learning rate of 1e-3. The model is trained for a total of 24 epochs while the learning rate is decayed by 10 at the 12-th epoch. In the fine stage, we train the regression model with Adam optimizer with a learning rate of 3e-4 for 12 epochs. 

\subsection{Evaluation Criteria} 
In alignment with the current research, we adopt retrieve recall at top $k$ (for $k = 1, 3, 5$) as our metric for evaluating the coarse stage capability of text-cell retrieving. For the fine stage evaluation, we use the normalized Euclidean distance error between the ground-truth position and the predicted position. Meanwhile, to evaluate the comprehensive effectiveness of our IFRP-T2P, we assess localization performance based on the top $k$ retrieval outcomes, where $k$ is set to 1, 5, and 10, and subsequently report on the localization recall. Localization recall is quantified as the fraction of text query accurately localized within predefined error margins, which is typically less than 5, 10, or 15 meters.

\subsection{Results}

\noindent\textbf{Pipeline localization.}
To conduct a comprehensive evaluation of IFRP-T2P, we compare our model with three state-of-the-art models \cite{Kolmet2022Text2PosTC, RET, xia2024text2loc} in two scenarios: one takes the ground-truth instance as input and the other uses the raw point cloud as input. In the context of raw point cloud scenario, we additionally compare our model with the 2D and 3D visual localization baselines,  specifically NetVlad \cite{Arandjelovic_Gronat_Torii_Pajdla_Sivic_2018} and PointNetVlad \cite{Uy2018PointNetVLADDP} to demonstrate the efficiency of our multi-modal model. We report the NetVlad result available in Text2Pos. For PointNetVlad, we employ a down-sampled one-third central region of the cell point cloud as the query to retrieve the top-$k$ cells and adopt the center of the cell as the predicated position, similar as the NetVlad setting in Text2Pos. Table~\ref{tab:fine_results} illustrates the top-$k$ $(k=1/5/10)$ recall rate of different error thresholds $\epsilon < 5/10/15 m$ for comparison. Our IFRP-T2P achieves $0.23/0.53/0.64$ at top-1/5/10 under the error bound $\epsilon < 5m$ on validation set. In the scenario where raw point cloud is adopted as input, our model outperforms the Text2Loc with prior instance segmentation model by $27\%/23\%/17\%$ at top-1/5/10 under the error bound $\epsilon < 5m$. Comparing with the Text2Loc, our model shows comparable performance—lagging by merely $17\%/8\%/7\%$ at top-1/5/10 under the error bound $\epsilon < 5m$. Note that we report the result of our reproduced Text2Loc and the values available in the original publication of RET and Text2Pos.  This trend is also observed in the test set, indicating our IFRP-T2P model effectively diminishes the dependency on ground-truth instances as input and effectively integrates relative position information in the coarse-to-fine localization process to improve the overall localization performance. Additionally, our model outperforms NetVlad and PointNetVlad by $28\%/82\%/88\%$ and $10\%/20\%/19\%$ at top-1/5/10 under the error bound $\epsilon < 5m$. This result indicates our multi-modal model is more effective comparing to the 2D/3D mono-modal models.

\textbf{Coarse text-cell retrieval.} In align with current works, we evaluate coarse text-cell retrieval performance on the KITTI360Pose validation and test set. Table~\ref{tab:coarse_results} presents the top-1/3/5 recall for each method. Our IFRP-T2P achieves the recall of $0.24/0.46/0.57$, on the validation set. In the scenario where raw point cloud is adopted as input, our model outperforms the Text2Loc with prior instance segmentation model by $26\%/24\%/23\%$ at top-1/3/5 recall rates. Comparing with the Text2Loc, our model shows comparable performance—lagging by merely $17\%/10\%/8\%$ at top-1/3/5 recall rates on the validation set. This trend is also observed in the test set, indicating that our proposed instance query extractor could effectively generate instance queries with semantic information and the RowColRPA module could capture the crucial relative position information of potential instances. More qualitative results are given in Section \ref{sec: vis}.

\textbf{Fine position estimation.} To evaluate the fine stage model performance, we use the paired cells and text descriptions as input and compute the normalized Euclidean distance between the ground-truth position and the predicted position. We normalize the side length of the cell to 1. Table~\ref{tab: regressor} shows that, under the instance-free scenario, our fine position estimation model yields $0.118$ normalized Euclidean distance error, which is $2\%$ and $9\%$ lower comparing to Text2Loc and Text2Pos in validation set. This trend is also observed in the test set. This result indicates that our RPCA module effectively incorporates the spatial relation information in the text and point cloud feature fusion process, and thus improves the fine position estimation performance.


\section{Performance analysis}
\label{sec: discussions}

\subsection{Ablation Study}
The following ablation studies evaluate the effectiveness of the relative position-aware components in the two stages.

\begin{table}[t]
    \centering
    \resizebox{0.47\textwidth}{!}{
    \begin{tblr}{
      cells = {c},
      cell{1}{1} = {r=3}{},
      cell{1}{2} = {c=6}{},
      cell{2}{2} = {c=3}{},
      cell{2}{5} = {c=3}{},
      vline{5} = {1-3}{solid}, 
      hline{1,8} = {-}{0.08em},
      hline{2-3} = {2-7}{0.03em},
      hline{4,7} = {-}{0.05em},
    }
    Methods      & Text-cell Retrieval Recall $\uparrow$ &               &               &               &               &               \\
                & Validation Set                     &               &               & Test Set      &               &               \\
                & $k=1$                              & $k=3$         & $k=5$         & $k=1$         & $k=3$         & $k=5$         \\
    Naive      & 0.19                               & 0.38          & 0.48          & 0.18          & 0.33          & 0.41          \\
    Value     & 0.22                               & 0.42          & 0.52          & 0.20          & 0.35          & 0.43          \\
    Row      & 0.22                               & 0.42          & 0.53          & 0.20          & 0.35          & 0.44          \\
    RowCol (Ours) & \textbf{0.24}                      & \textbf{0.46} & \textbf{0.57} & \textbf{0.23} & \textbf{0.39} & \textbf{0.48} 
    \end{tblr}}
    \caption{Ablation study of the row-column relative position-aware self-attention (RowColRPA) in coarse stage. ``Naive'' refers to the employment of standard self-attention mechanisms. ``Value'' denotes integrating pooled relative position features into the value, akin to the methodology described in RET~\cite{RET}. ``Row'' signifies adding the query with row-wise pooled relative position features. ``RowCol'' stands for our proposed RowColRPA.
    }
    \label{tab: ablation study}
    \vspace{-0.5em}
\end{table}

\begin{table}[t]
    \centering
    \resizebox{0.47\textwidth}{!}{%
    \begin{tblr}{
      cells = {c},
      cell{1}{2} = {c=6}{},
      cell{2}{2} = {c=3}{},
      cell{2}{5} = {c=3}{},
      vline{5} = {1-3}{solid}, 
      hline{1,6} = {-}{0.08em},
      hline{2-3} = {2-7}{0.03em},
      hline{4} = {-}{0.05em},
    }
                    & Localization Recall ($\epsilon < 5m$) $\uparrow$ &               &               &               &               &               \\
    Methods         & Validation Set                                   &               &               & Test Set      &               &               \\
                    & $k=1$                                            & $k=5$         & $k=10$        & $k=1$         & $k=5$         & $k=10$        \\
    Naive       & 0.20                                             & 0.48          & 0.59          & 0.19          & 0.43          & 0.52          \\
    RPCA (Ours) & \textbf{0.23}                                    & \textbf{0.53} & \textbf{0.64} & \textbf{0.22} & \textbf{0.47} & \textbf{0.58} 
    \end{tblr}}
    \caption{Ablation study of the relative position-aware cross-attention (RPCA) in fine stage. ``Naive'' indicates the application of standard cross-attention in the multi-modal fusion module. }
    \label{tab: fine ablation study}
    \vspace{-0.5em}
\end{table}

\label{sec: ablation}
\noindent{\bf RowColRPA.}
To evaluate the effectiveness of RowColRPA in the coarse stage, we compare it with different variants, as shown in Table~\ref{tab: ablation study}. The result reveals that incorporating a relative position attribute into the value component yields a modest enhancement of $15\%/10\%/8\%$ at the top-1/3/5 recall metrics, respectively, when compared to the conventional self-attention mechanism. Incorporating the pooled relative position feature into the query results in nearly the same level of improvement, with a marginally higher increase observed at the top-5 recall rate. In contrast, the novel strategy of integrating a row-wise pooled relative position feature with the query, and introducing a column-wise pooled relative position feature to the key, results in a significant performance boost of $26\%/21\%/18\%$ against the standard self-attention at the top-1/3/5 recall benchmarks on the validation dataset. This demonstrates the pronounced superiority and efficiency of the proposed RowColRPA in capturing spatial relationships and enhancing retrieval performance.

\noindent{\bf RPCA.}
To analyse the effectiveness of RPCA in the fine stage, we compare it with the variant using standard cross-attention, as shown in Table~\ref{tab: fine ablation study}. The result shows that our RPCA leads to $15\%/10\%/8\%$ improvement comparing to the standard self-attention at top-1/5/10 localization recall rates, respectively. It demonstrates the capability of RPCA to effectively integrate relative position information during the multi-modal fusion process, thereby enhancing localization accuracy.

\begin{figure*}[h]
    \centering
    \includegraphics[width=0.995\textwidth]{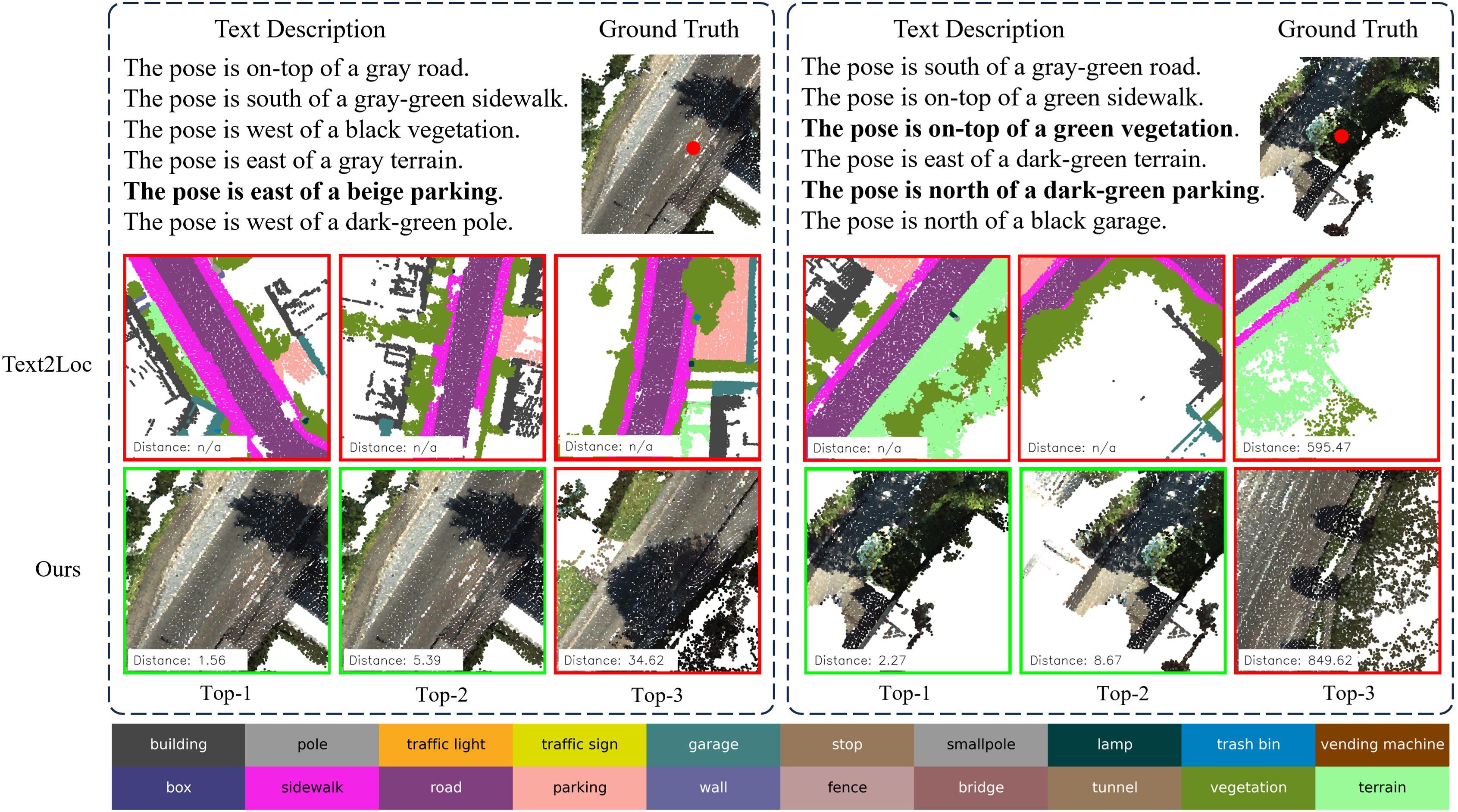}
    \vspace{-0.5em}
    \caption{Comparison of the top-3 retrieved cells between Text2Loc \cite{xia2024text2loc} and IFRP-T2P. The numbers within the top-3 retrieval submaps denote the center distances between the retrieved submaps and the ground-truth, with ``n/a'' indicating distances exceeding 1000 meters. Green boxes highlight the positive submaps, which contain the target location, whereas red boxes delineate the negative submaps that do not contain the target. 
    }
    \label{fig:qualit_results}
    \vspace{-0.5em}

\end{figure*}

\subsection{Qualitative Analysis}
\label{sec: vis}
In addition to the quantitative metrics, we also offer a qualitative analysis comparing the top-1/2/3 retrieved cells by Text2Loc \cite{xia2024text2loc} and IFRP-T2P, as depicted in Fig.~\ref{fig:qualit_results}. In the first column, the result indicates that both models can retrieve cells with the described instances. However, there are notable differences in their accuracy with respect to the spatial relation descriptions provided. Specifically, for the ``beige parking'' instance, which is described as being located to the west of the cell, the retrieval result of Text2Loc inaccurately places it to the e   ast of the cell centers. Conversely, IFRP-T2P correctly locates this instance to the east of the center, aligning with the given description. In the second column, the text hints describe that the pose is on-top of a ``dark-green vegetation'' and is north of a ``dark-green parking''. For Text2Loc, the parking is found to the north of the cell center in the top-1/2 retrieved cells, and the vegetation is located at the margin area of the top-1/2/3 retrieved cells, discrepant from the text description. For IFRP-T2P, however, the parking appears on the south of the cell center in the top-1/2 retrieved cells, and the vegetation appears on the center of the top-1/2/3 retrieved cells, which matches with the text description. Notably, in both cases, only the third retrieved cell by IFRP-T2P exceeds the error threshold. This evidence solidifies the superior capacity of IFRP-T2P to interpret and utilize relative position information in comparison to Text2Loc. More case studies of our IFRP-T2P are provided in the supplement material.

\subsection{Text Embedding Analysis}

Recent years have seen the emergence of large language models (encoders) like BERT \cite{Devlin2019BERTPO}, RoBERTa \cite{Liu2019RoBERTaAR}, T5 \cite{2020t5}, and the CLIP \cite{Radford2021LearningTV} text encoder, each is trained with varied tasks and datasets. Text2Loc highlights that a pre-trained T5 model significantly enhances text and point cloud feature alignment. Yet, the potential of other models, such as RoBERTa and the CLIP text encoder, known for their excellence in visual grounding tasks, is not explored in their study. Thus, we conduct a comparative analysis of T5-small, RoBERTa-base, and the CLIP text encoder within our model framework. The result in Table~\ref{tab: text_enc} indicates that the T5-small (61M) achieves $0.24/0.46/0.57$ at the top-1/3/5 recall metrics, incrementally outperforming RoBERTa-base (125M) and CLIP text (123M) with fewer parameters. 

\begin{table}[t]
    \centering
    \small
    \resizebox{0.475\textwidth}{!}{
    \begin{tblr}{
      colspec={X[1,c]X[1,c]X[1,c]X[1,c]},
      cells = {c},
      cell{1}{1} = {r=2}{},
      cell{1}{2} = {c=3}{},
      hline{1,6} = {-}{0.08em},
      hline{2} = {2-7}{0.03em},
      hline{3} = {1-7}{0.03em},
      hline{5} = {-}{0.05em},
    }
    Text encoder      & Text-cell Retrieval Recall $\uparrow$ (Validation)    \\
                & $k=1$                              & $k=3$         & $k=5$         \\
    RoBERTa \cite{Liu2019RoBERTaAR}      &     0.23                      &     0.43       &     0.53      \\
    CLIP text \cite{Radford2021LearningTV}     &         0.23                      &     0.45       &    0.56      \\ 
    T5 \cite{2020t5} & \textbf{0.24}                      & \textbf{0.46} & \textbf{0.57}  
    \end{tblr}
    }
    \caption{Comparison of different text encoders.}
    \label{tab: text_enc}
    \vspace{-2em}
\end{table}

\section{Conclusion}
In this paper, we propose a IFRP-T2P model for 3D point cloud localization based on a few natural language descriptions, which is the first approach to directly take raw point clouds as input, eliminating the need for ground-truth instances.  Additionally, we propose the RowColRPA in the coarse stage and the RPCA in the fine stage to fully leverage the spatial relation information. Through extensive experiments, IFRP-T2P achieves comparable performance with the state-of-the-art model, Text2Loc, which relies on ground-truth instances. Moreover, it surpasses the Text2Loc using instance segmentation model as prior. Our approach expands the usability of existing text-to-point-cloud localization models, enabling their application in scenarios where few instance information is available.  Our future work will focus on applying IFRP-T2P for navigation in real-world robotic applications, bridging the gap between theoretical models and practical utility in autonomous navigation and interaction.

\newpage
\bibliographystyle{ACM-Reference-Format}
\balance
\bibliography{sample-base}

\clearpage
\maketitlesupplementary

\section{Details of KITTI360Pose Dataset}

\begin{figure}[h]
    \begin{center}
    \includegraphics[width=0.475\textwidth]{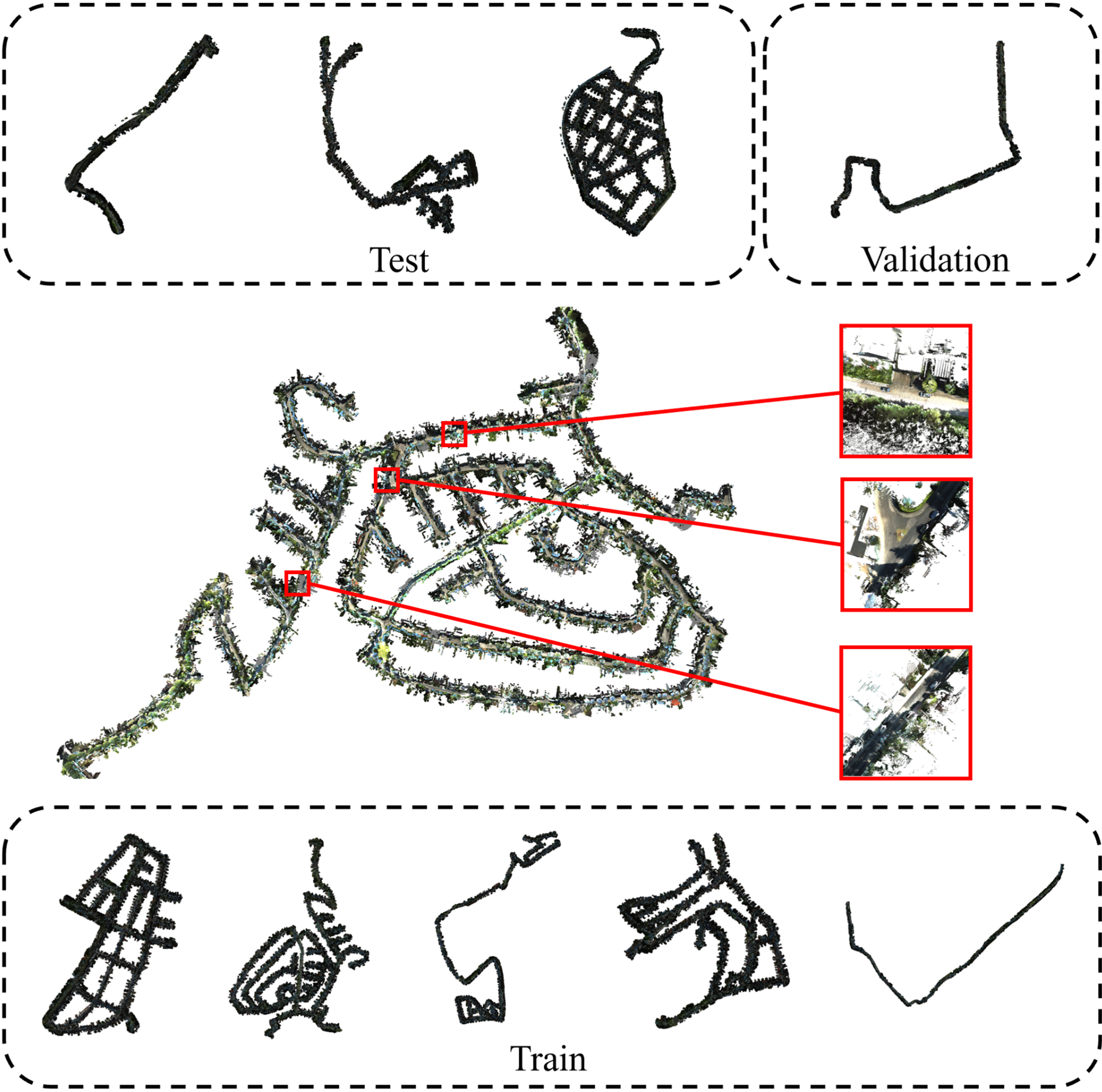}
    \end{center}
    \vspace{-1em}
    \caption{Visualization of the KITTI360Pose dataset. The trajectories of five training
sets, three test sets, and one validation set are shown in the dashed borders. One colored point cloud scene and three cells are shown in the middle.}
    \label{fig:scenes}
\end{figure}

For the KITTI360Pose dataset, we show the trajectories of training set ($11.59 km^2$), validation set ($1.78 km^2$), and test set ($2.14 km^2$) in Fig.~\ref{fig:scenes}. To generate position and text description pairs, equidistant points along the map trajectories are selected. Around each point, 8 additional positions are randomly selected to enrich the dataset. Positions with fewer than 6 nearby instances are excluded. For each position, text descriptions of nearby objects are generated using a predefined sentence template: ``The pose is on [direction] of a [color] [instance].''
\vspace{-1em}
\vspace{-0.5em}

\section{More Experiments on the Instance Query Extractor}

\begin{table}[h]
    \centering
    \resizebox{0.475\textwidth}{!}{
    \begin{tblr}{
      cells = {c},
      cell{1}{1} = {r=3}{},
      cell{1}{2} = {c=6}{},
      cell{2}{2} = {c=3}{},
      cell{2}{5} = {c=3}{},
      vline{5} = {1-3}{solid}, 
      hline{1,7} = {-}{0.08em},
      hline{2-3} = {2-7}{0.03em},
      hline{4,6} = {-}{0.05em},
    }
    Query Number      & Localization Recall ($\epsilon < 5m$) $\uparrow$ &               &               &               &               &               \\
                & Validation Set                     &               &               & Test Set      &               &               \\
                & $k=1$                              & $k=5$         & $k=10$         & $k=1$         & $k=5$         & $k=10$         \\
    16      & 0.20                              & 0.47          & 0.58          & 0.18          & 0.41          & 0.51          \\
    32     & 0.21                               & 0.50         & 0.61          & 0.20          & 0.44         & 0.55          \\
    24 (Ours) & \textbf{0.23}                      & \textbf{0.53} & \textbf{0.64} & \textbf{0.22} & \textbf{0.47} & \textbf{0.58} 
    \end{tblr}}
    \caption{Ablation study of the query number on KITTI360Pose dataset.}
    \label{tab: query}
\end{table}

We conduct an additional experiment to assess the impact of the number of queries on the performance of our instance query extractor. As detailed in Table \ref{tab: query}, we evaluate the localization recall rate using 16, 24, and 32 queries. The result demonstrates that using 24 queries yields the highest localization recall rate, i.e, $0.23/0.53/0.64$ on the validation set and $0.22/0.47/0.58$ on the test set. This finding suggests that the optimal number of queries for maximizing the effectiveness of our model is 24.

\section{Text-cell Embedding Space Analysis}
\begin{figure}[htb]
    \begin{center}
    \includegraphics[width=0.475\textwidth]{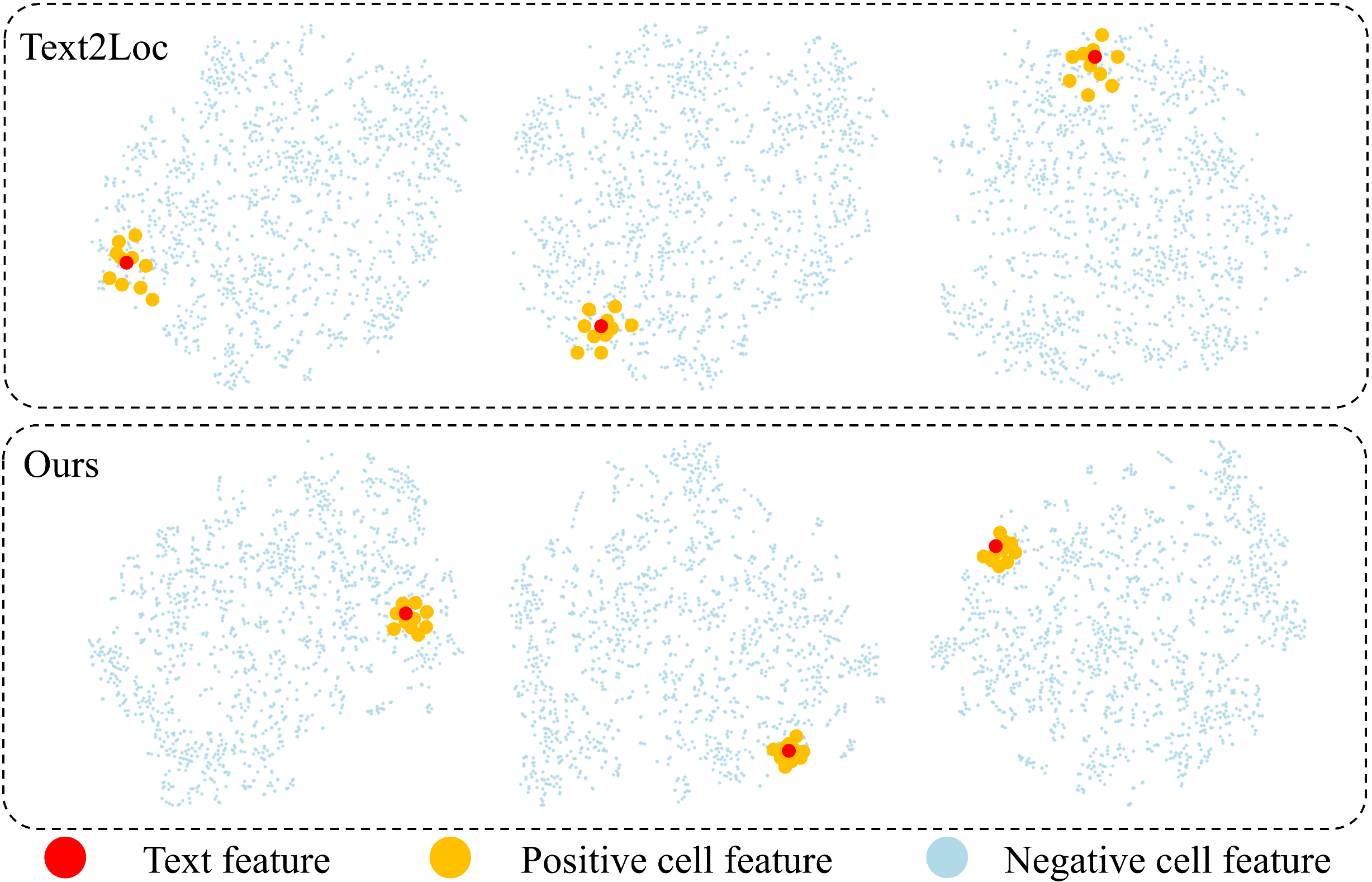}
    \end{center}
    \vspace{-1em}
    \caption{T-SNE visualization for the text features and cell features in the coarse stage.}
    \label{fig:embeddings}
\end{figure}

Fig.~\ref{fig:embeddings} shows the aligned text-cell embedding space via T-SNE \cite{van2008visualizing}. Under the instance-free scenario, we compare our model with Text2loc \cite{xia2024text2loc} using a pre-trained instance segmentation model, Mask3D \cite{Schult23ICRA}, as a prior step. It can be observed that Text2Loc results in a less discriminative space, where positive cells are relatively far from the text query feature. In contrast, our IFRP-T2P effectively reduces the distance between positive cell features and text query features within the embedding space, thereby creating a more informative embedding space. This enhancement in the embedding space is critical for improving the accuracy of text-cell retrieval.
\vspace{-0.5em}

\begin{figure*}[htb]
    \begin{center}
    \includegraphics[width=0.99\textwidth]{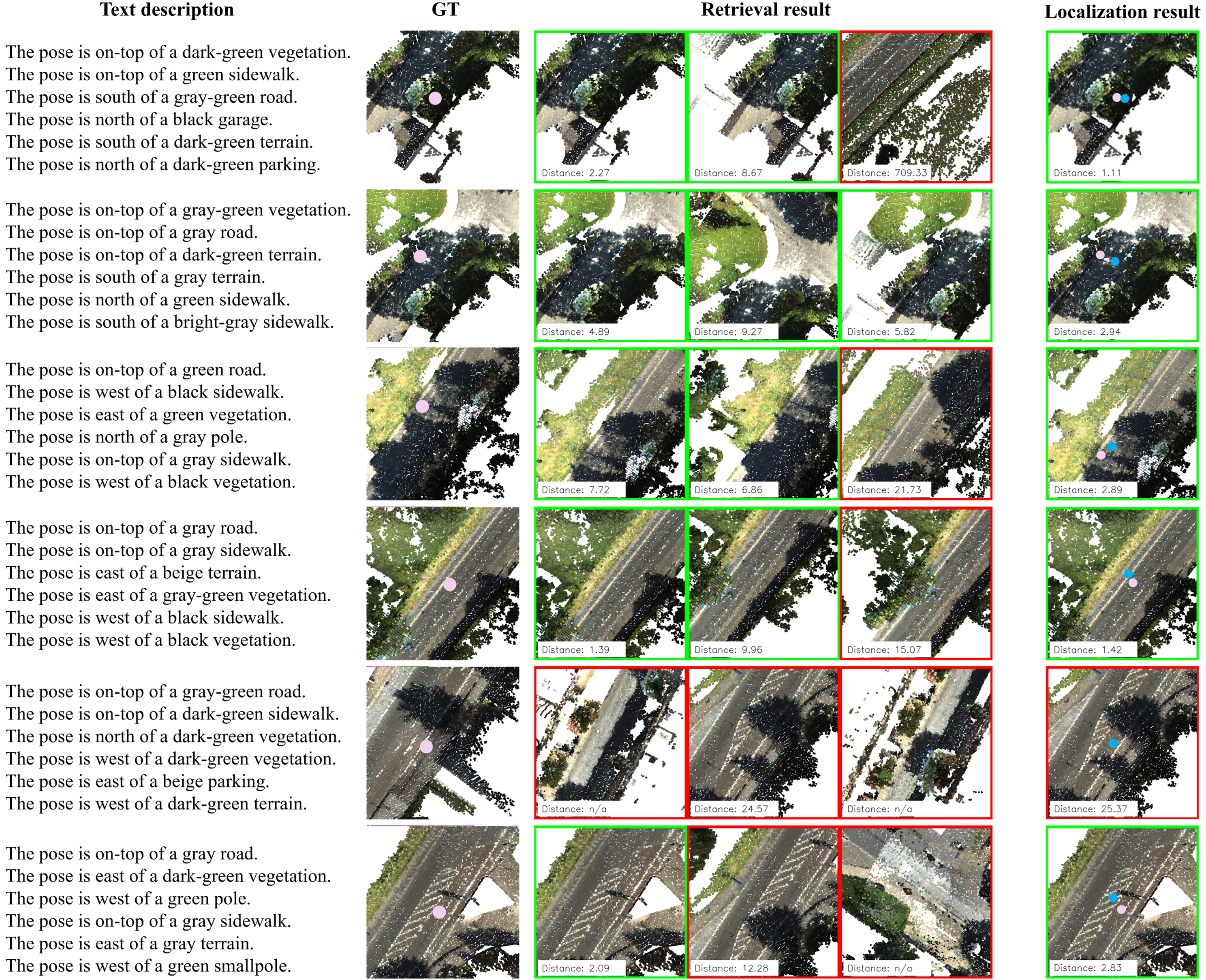}
    \end{center}
    \vspace{-1em}
    \caption{Localization results on the KITTI360Pose dataset. In the coarse stage, the numbers in the top 3 retrieval submaps represent the center distances between retrieved submaps and the ground truth. For fine localization, pink and blue points represent the ground-truth localization and the predicted location, with the number indicating the distance between them.}
    \label{fig:vis}
\end{figure*}

\section{More visualization Results}
Fig.~\ref{fig:vis} shows more visualization results including both the retrieval outcomes and the results of fine position estimation. The results suggest that the coarse text-cell retrieval serves as a foundational step in the overall localization process. The subsequent fine position estimation generally improves the localization performance. However, there are cases where the accuracy of this fine estimation is compromised, particularly when the input descriptions are vague. This detrimental effect on accuracy is illustrated in the 4-th row and 6-th row if Fig.~\ref{fig:vis}.

\section{Point Cloud Robustness Analysis}
\begin{figure}[htb]
    \begin{center}
    \includegraphics[width=0.47\textwidth]{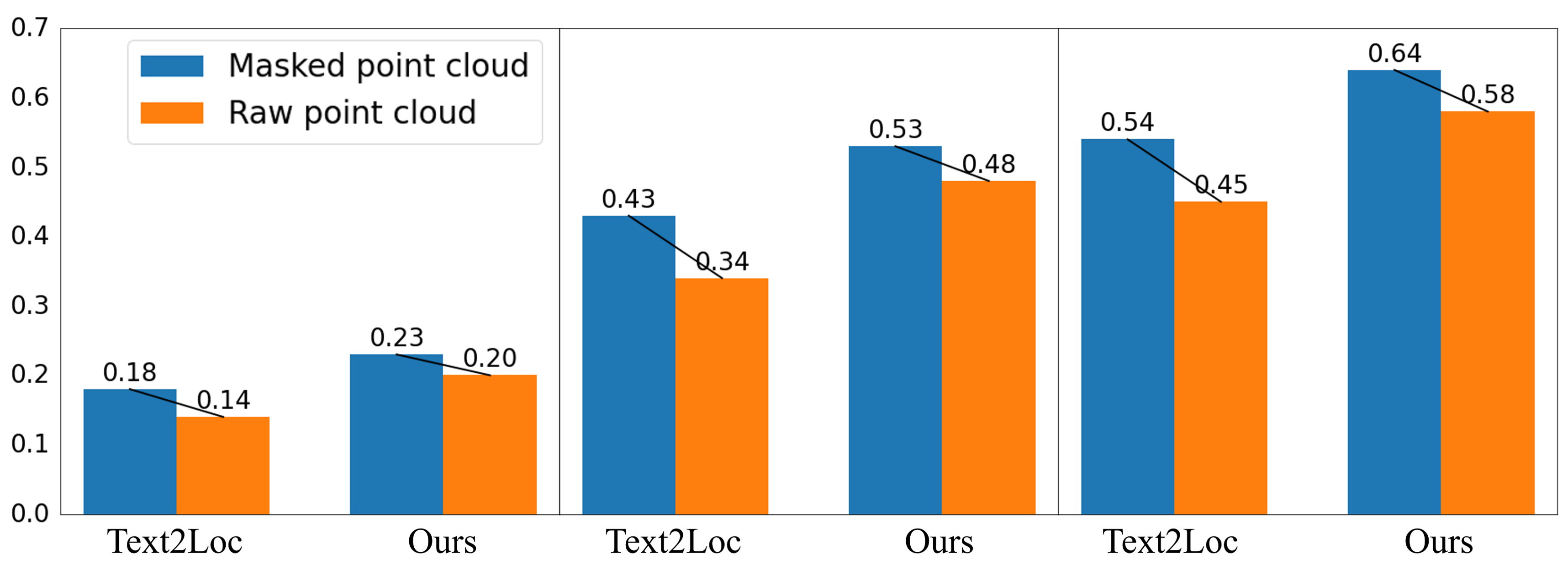}
    \end{center}
    \vspace{-0.5em}
    \caption{Point cloud robustness analysis.}
    \label{fig: pc_robust}
\end{figure}

Previous works \cite{xia2024text2loc, RET, Kolmet2022Text2PosTC} focused solely on examining the impact of textual modifications on localization accuracy, ignoring the impact of point cloud modification. In this study, we further consider the effects of point cloud degradation, which is crucial for fully analyzing our IFRP-T2P model. Unlike the accumulated point clouds provided in the KITTI360Pose dataset, LiDAR sensors typically capture only sparse point clouds in real-world settings. To assess the robustness of our model under conditions of point cloud sparsity, we conduct experiments by randomly masking out one-third of the points and compare these results to those obtained using raw point clouds. As illustrated in Fig.~\ref{fig: pc_robust}, when taking the masked point cloud as input, our IFRP-T2P model achieves a localization recall of $0.20$ at top-$1$ with an error bound of $\epsilon < 5m$ on the validation set. Compared to Text2Loc, which shows a degradation of $22.2\%$, our model exhibits a lower degradation rate of $15\%$. This result indicates that our model is more robust to point cloud variation.

\end{document}


\title{Supplementary Materials: Instance-free Text to Point Cloud Localization with Relative Position Awareness}


\author{Anonymous Authors}

\maketitle

\section{Details of KITTI360Pose Dataset}

\begin{figure}[htb]
    \begin{center}
    \includegraphics[width=0.475\textwidth]{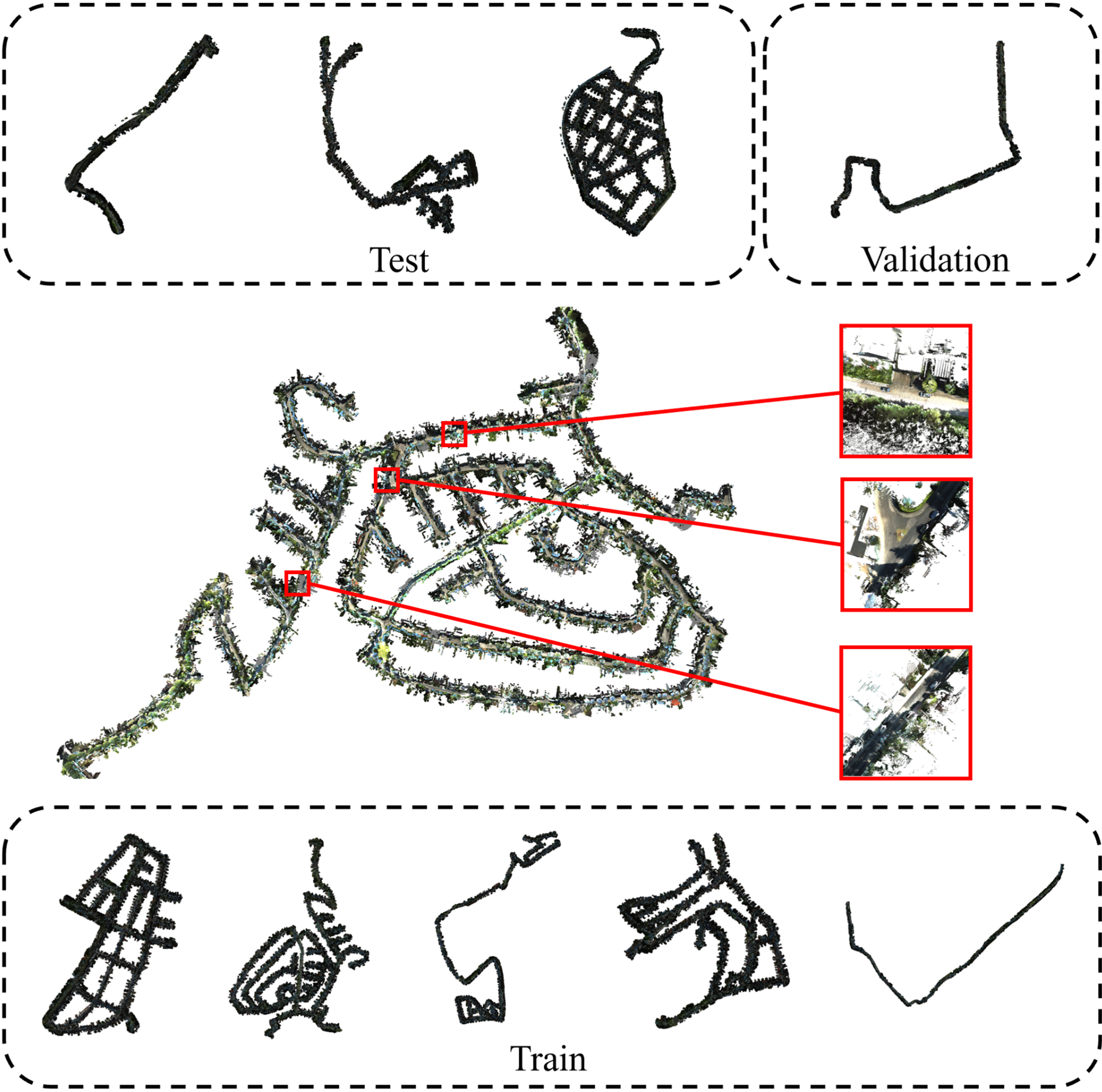}
    \end{center}
    \vspace{-1em}
    \caption{Visualization of the KITTI360Pose dataset. The trajectories of five training
sets, three test sets, and one validation set are shown in the dashed borders. One colored point cloud scene and three cells are shown in the middle.}
    \label{fig:scenes}
\end{figure}

For the KITTI360Pose dataset, we show the trajectories of training set ($11.59 km^2$), validation set ($1.78 km^2$), and test set ($2.14 km^2$) in Fig.~\ref{fig:scenes}.
To generate position and text description pairs, equidistant points along the map trajectories are selected. Around each point, 8 additional positions are randomly selected to enrich the dataset. Positions with fewer than 6 nearby instances are excluded. For each position, text descriptions of nearby objects are generated using a predefined sentence template: ``The pose is on [direction] of a [color] [instance].''
\vspace{-1.5em}
\vspace{-0.5em}

\section{More Experiments on the Instance Query Extractor}
We conduct an additional experiment to assess the impact of the number of queries on the performance of our instance query extractor. As detailed in Table \ref{tab: query}, we evaluate the localization recall rate using 16, 24, and 32 queries. The result demonstrates that using 24 queries yields the highest localization recall rate, i.e, $0.23/0.53/0.64$ on the validation set and $0.22/0.47/0.58$ on the test set. This finding suggests that the optimal number of queries for maximizing the effectiveness of our model is 24.

\begin{table}[h]
    \centering
    \resizebox{0.475\textwidth}{!}{
    \begin{tblr}{
      cells = {c},
      cell{1}{1} = {r=3}{},
      cell{1}{2} = {c=6}{},
      cell{2}{2} = {c=3}{},
      cell{2}{5} = {c=3}{},
      vline{5} = {1-3}{solid}, 
      hline{1,7} = {-}{0.08em},
      hline{2-3} = {2-7}{0.03em},
      hline{4,6} = {-}{0.05em},
    }
    Query Number      & Localization Recall ($\epsilon < 5m$) $\uparrow$ &               &               &               &               &               \\
                & Validation Set                     &               &               & Test Set      &               &               \\
                & $k=1$                              & $k=5$         & $k=10$         & $k=1$         & $k=5$         & $k=10$         \\
    16      & 0.20                              & 0.47          & 0.58          & 0.18          & 0.41          & 0.51          \\
    32     & 0.21                               & 0.50         & 0.61          & 0.20          & 0.44         & 0.55          \\
    24 (Ours) & \textbf{0.23}                      & \textbf{0.53} & \textbf{0.64} & \textbf{0.22} & \textbf{0.47} & \textbf{0.58} 
    \end{tblr}}
    \caption{Ablation study of the query number on KITTI360Pose dataset.}
    \label{tab: query}
    \vspace{-1.2em}
\end{table}



\section{Text-cell Embedding Space Analysis}
\begin{figure}[htb]
    \begin{center}
    \includegraphics[width=0.475\textwidth]{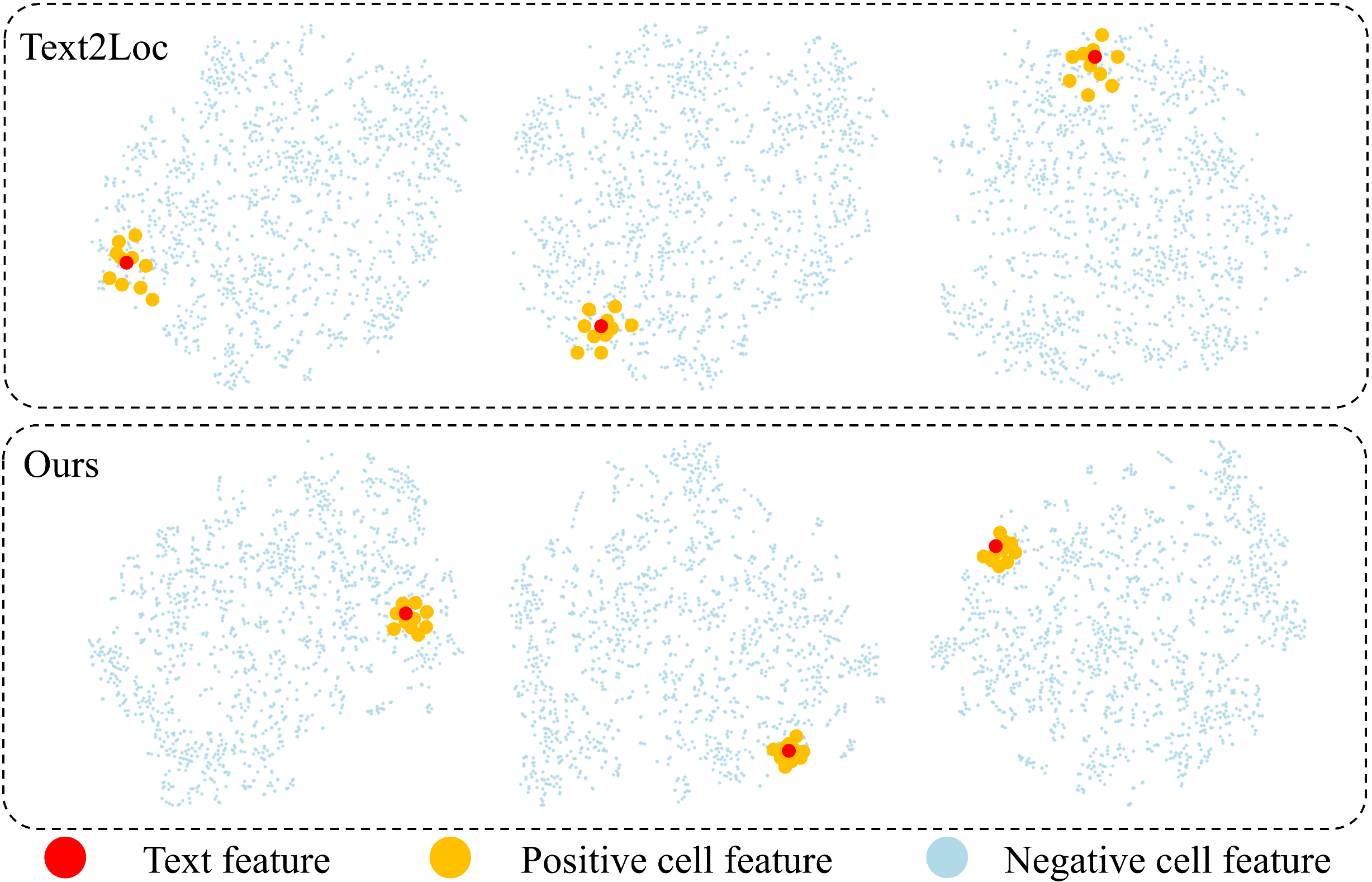}
    \end{center}
    \vspace{-1em}
    \caption{T-SNE visualization for the text features and cell features in the coarse stage.}
    \label{fig:embeddings}
\end{figure}

Fig.~\ref{fig:embeddings} shows the aligned text-cell embedding space via T-SNE \cite{van2008visualizing}. Under the instance-free scenario, we compare our model with Text2loc \cite{xia2024text2loc} using a pre-trained instance segmentation model, Mask3D \cite{Schult23ICRA}, as a prior step. It can be observed that Text2Loc results in a less discriminative space, where positive cells are relatively far from the text query feature. In contrast, our IFRP-T2P effectively reduces the distance between positive cell features and text query features within the embedding space, thereby creating a more informative embedding space. This enhancement in the embedding space is critical for improving the accuracy of text-cell retrieval.
\vspace{-1em}
\vspace{-1em}

\begin{figure*}[htb]
    \begin{center}
    \includegraphics[width=0.99\textwidth]{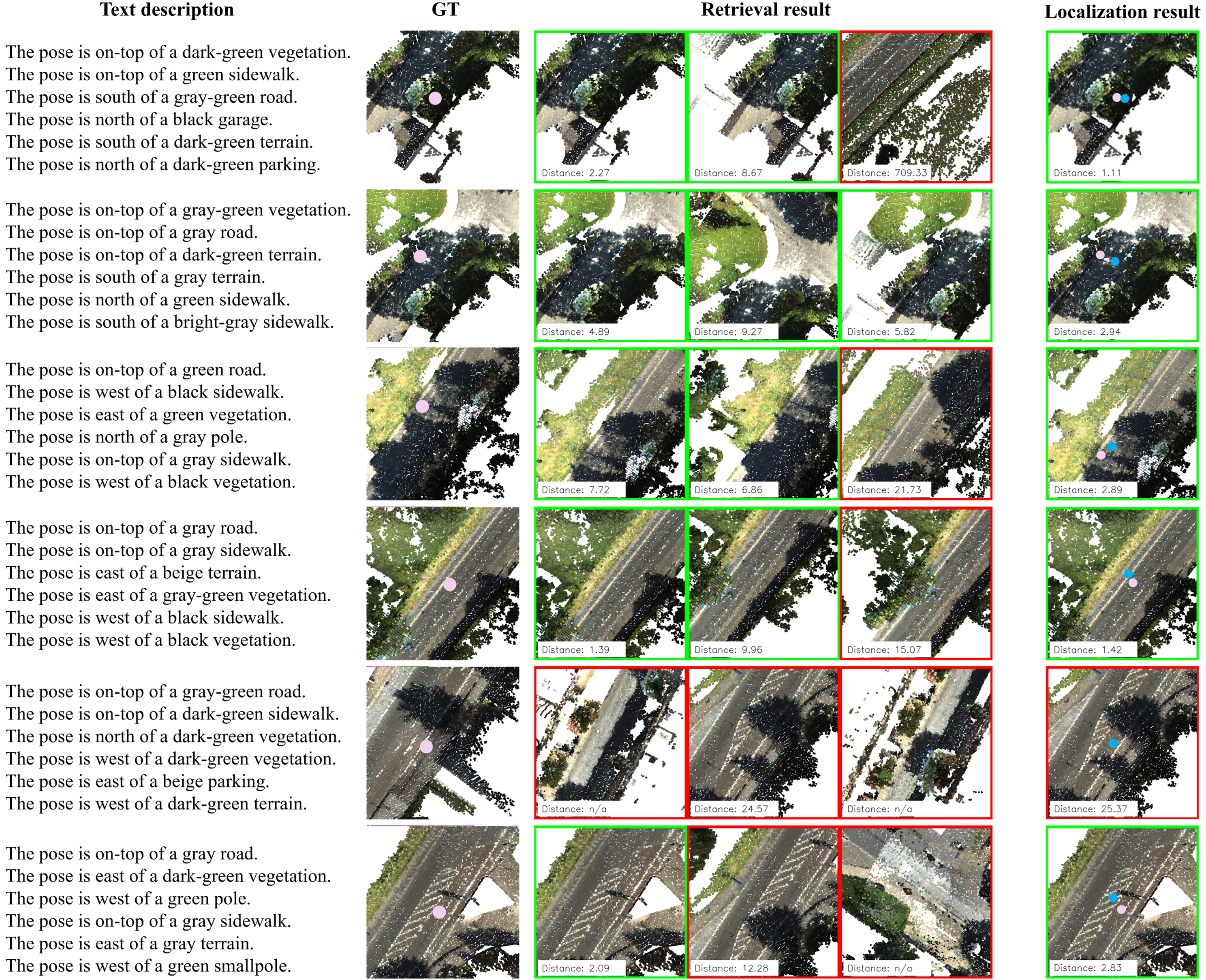}
    \end{center}
    \vspace{-1em}
    \caption{Localization results on the KITTI360Pose dataset. In the coarse stage, the numbers in the top 3 retrieval submaps represent the center distances between retrieved submaps and the ground truth. For fine localization, pink and blue points represent the ground-truth localization and the predicted location, with the number indicating the distance between them.}
    \label{fig:vis}
\end{figure*}

\section{More visualization Results}
Fig.~\ref{fig:vis} shows more visualization results including both the retrieval outcomes and the results of fine position estimation. The results suggest that the coarse text-cell retrieval serves as a foundational step in the overall localization process. The subsequent fine position estimation generally improves the localization performance. However, there are cases where the accuracy of this fine estimation is compromised, particularly when the input descriptions are vague. This detrimental effect on accuracy is illustrated in the 4-th row and 6-th row if Fig.~\ref{fig:vis}.

\section{Point Cloud Robustness Analysis}
\begin{figure}[htb]
    \begin{center}
    \includegraphics[width=0.47\textwidth]{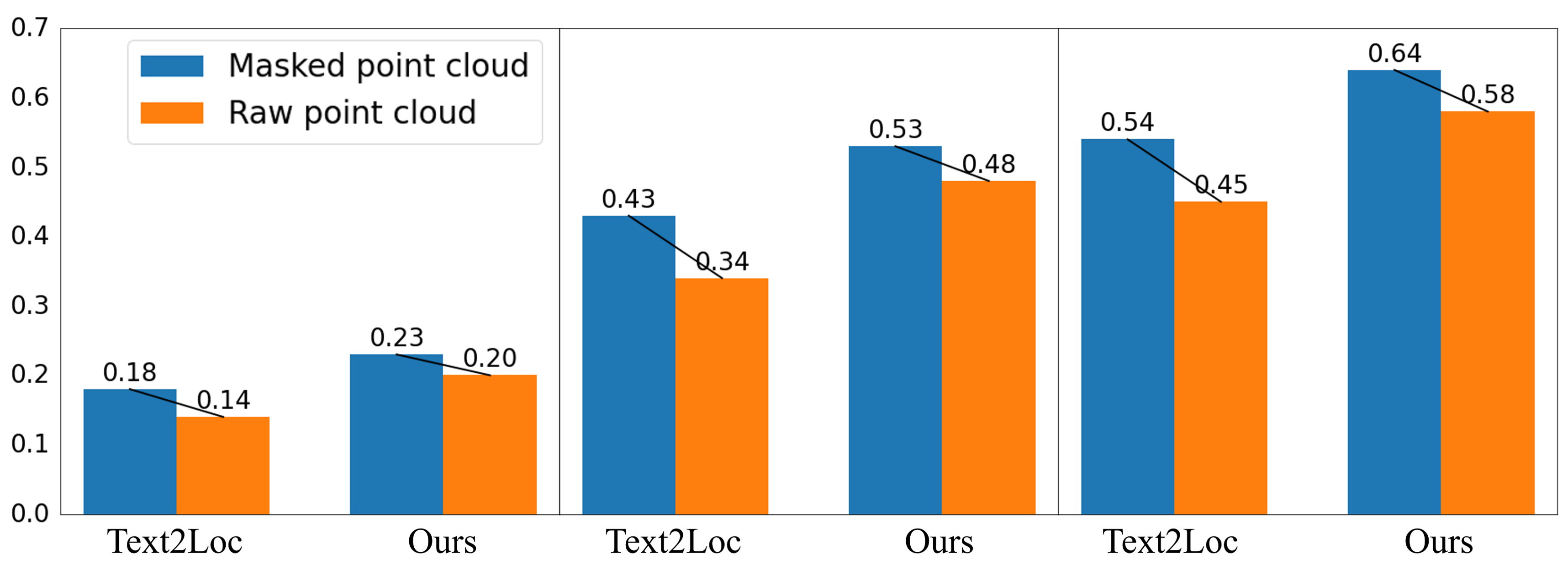}
    \end{center}
    \vspace{-0.5em}
    \caption{Point cloud robustness analysis.}
    \label{fig: pc_robust}
\end{figure}

Previous works \cite{xia2024text2loc, RET, Kolmet2022Text2PosTC} focused solely on examining the impact of textual modifications on localization accuracy, ignoring the impact of point cloud modification. In this study, we further consider the effects of point cloud degradation, which is crucial for fully analyzing our IFRP-T2P model. Unlike the accumulated point clouds provided in the KITTI360Pose dataset, LiDAR sensors typically capture only sparse point clouds in real-world settings. To assess the robustness of our model under conditions of point cloud sparsity, we conduct experiments by randomly masking out one-third of the points and compare these results to those obtained using raw point clouds. 
As illustrated in Fig.~\ref{fig: pc_robust}, when taking the masked point cloud as input, our IFRP-T2P model achieves a localization recall of $0.20$ at top-$1$ with an error bound of $\epsilon < 5m$ on the validation set. Compared to Text2Loc, which shows a degradation of $22.2\%$, our model exhibits a lower degradation rate of $15\%$. This result indicates that our model is more robust to point cloud variation.

\newpage
\bibliographystyle{ACM-Reference-Format}
\balance
\bibliography{sample-base}
